%% file: physics.tex
\documentclass[lettersize,journal]{IEEEtran}
\usepackage{algorithm}
\usepackage{algpseudocode}
\usepackage{graphicx}
\usepackage{textcomp}
\usepackage{xcolor}
\usepackage{balance}
\usepackage{xspace}
\usepackage{enumitem}
\usepackage{hyperref}
\usepackage[many]{tcolorbox}
\usepackage{subfigure}
\usepackage{booktabs}
\usepackage[normalem]{ulem}
\usepackage{colortbl}
\usepackage{siunitx}
\useunder{\uline}{\ul}{}
\usepackage[misc]{ifsym}
\usepackage{threeparttable}
\usepackage{balance}
\usepackage{subcaption}

\makeatletter
\def\namedlabel#1#2{\begingroup
	#2%
	\def\@currentlabel{#2}%
	\phantomsection\label{#1}\endgroup
}
\makeatother

\newcommand{\physicsbug}{physics failure\xspace}
\newcommand{\physicsbugs}{physics failures\xspace}

\newcommand{\physicsbugsCap}{Physics Failures\xspace}
\newcommand{\physicsbench}{\textsc{PhysiXFails}\xspace}
\newcommand{\physicsbenchmulti}{\textsc{PhysiXMultiFails}\xspace}

\newcommand{\numberOfPhysicsBugCategories}{17\xspace}

\newtcolorbox{answerbox}{%
	left=3pt,
	right=3pt,
	top=0pt,
	bottom=0pt,
	boxrule=0mm,
	colback=black!10!white,
	breakable,
	frame empty
}%

\newcommand{\todoaftersub}[1]{}

\definecolor{applegreen}{rgb}{0.55, 0.80, 0.4}

\let\diff\undefined %

\ifdefined\diff

\else

\fi

\definecolor{mygray}{gray}{.9}

\def\BibTeX{{\rm B\kern-.05em{\sc i\kern-.025em b}\kern-.08em
    T\kern-.1667em\lower.7ex\hbox{E}\kern-.125emX}}
\begin{document}

\title{Runtime Failure Hunting for Physics Engine Based Software Systems: How Far Can We Go?}

\author{Shuqing~Li,        
	Qiang~Chen,
	Xiaoxue~Ren,
	Michael~R.~Lyu
\IEEEcompsocitemizethanks{\IEEEcompsocthanksitem Shuqing~Li, and Michael~R.~Lyu are with The Chinese University of Hong Kong, China. E-mail:\{sqli21, lyu\}@cse.cuhk.edu.hk
\IEEEcompsocthanksitem Qiang~Chen, and Xiaoxue Ren are with the State Key Laboratory of Blockchain and Data
Security, Zhejiang University, Hangzhou, China. E-mail: \{qiangq688, xxren\}@zju.edu.cn
\IEEEcompsocthanksitem Xiaoxue Ren is the corresponding author.
}
}

\markboth{Journal of \LaTeX\ Class Files,~Vol.~14, No.~8, August~2021}%
{Shell \MakeLowercase{\textit{et al.}}: A Sample Article Using IEEEtran.cls for IEEE Journals}

\IEEEtitleabstractindextext{%

\input{./sections/abstract.tex}

}

\maketitle

\IEEEdisplaynontitleabstractindextext

\IEEEpeerreviewmaketitle

\input{./sections/introduction.tex}
\input{./sections/background.tex}

\input{./sections/benchmark.tex}

\input{./sections/taxonomy.tex}

\input{./sections/experiment.tex}

\input{sections/results_developer_study}

\input{./sections/discussion.tex}
\input{./sections/related-work.tex}
\input{./sections/conclusion.tex}

\appendix

\input{sections/developer_study}

\balance
\bibliographystyle{IEEEtran}
\bibliography{physics}

\end{document}

%% file: sections/abstract.tex
\begin{abstract}

Physics Engines (PEs) are fundamental software frameworks that simulate physical interactions in applications ranging from entertainment to safety-critical systems. Despite their importance, PEs suffer from \textit{\physicsbugs}, deviations from expected physical behaviors that can compromise software reliability, degrade user experience, and potentially cause critical failures in autonomous vehicles or medical robotics. Current testing approaches for PE-based software are inadequate, typically requiring white-box access and focusing on crash detection rather than semantically complex \physicsbugs. This paper presents the first large-scale empirical study characterizing \physicsbugs in PE-based software. We investigate three research questions addressing the manifestations of \physicsbugs, the effectiveness of detection techniques,
and developer perceptions of current detection practices. 
Our contributions include: (1) a taxonomy of \physicsbug manifestations; (2) a comprehensive evaluation of detection methods including deep learning, prompt-based techniques, and large multimodal models; 
and (3) actionable insights from developer experiences for improving detection approaches. To support future research, we release \physicsbench, code, and other materials at \href{https://sites.google.com/view/physics-failure-detection}{\color{blue}{https://sites.google.com/view/physics-failure-detection}}.
\end{abstract}

%% file: sections/introduction.tex
\section{Introduction}\todoaftersub{Experiments on Undetected bugs?}

\IEEEPARstart{P}{hysics}
engines (PEs) are software frameworks that provide digital simulations of physical interactions by modeling real-world phenomena, such as rigid body dynamics, collision detection, soft body deformation, and fluid dynamics~\cite{hussain2020unity, templet2021game, templet2021game, cuttingedgephysicsengines}. These engines have become essential in modern software development, especially in fields requiring realistic interactions and high-fidelity environments, including autonomous systems (e.g., automated vehicles and robotics), 3D interactive software (e.g., virtual or mixed reality in medical surgery assistance), scientific simulations, and entertainment software (e.g., games)~\cite{shah2018airsim, maciel2009using, DBLP:journals/tciaig/RenzMSW16}.
By providing high-fidelity simulations of physical phenomena, PEs empower developers to create immersive and interactive environments that closely emulate real-world conditions, significantly enhancing both user experience and software functionality.

Despite their importance, PEs face substantial challenges due to the complexity of accurately simulating physical laws. A common issue is the occurrence of \emph{\textbf{\physicsbugsCap}}, defined as deviations from expected physical behaviors within software systems that rely on PEs~\cite{paper:conf/kbse/XiaoLW23/phyfu}. These failures can significantly affect software reliability, leading to consequences ranging from degraded user experiences and cybersickness to critical failures in safety-sensitive systems like autonomous vehicles and surgical robotics~\cite{hussain2018autonomous}. However, detecting and diagnosing these failures remains challenging due to the intricate interplay between software code and underlying physics simulations.

Despite the significance of \physicsbugs, existing testing approaches for PE-based software are insufficient. Current methods primarily focus on white-box techniques that require access to the internal mechanics of the engines~\cite{he2017taming, liang2018fuzzing}, predominantly targeting crash detection while failing to address semantically complex \physicsbugs. Recent advances such as PhyFu~\cite{paper:conf/kbse/XiaoLW23/phyfu} have demonstrated promising approaches to fuzz testing PEs at the logic level, but these efforts are limited by their reliance on traditional fuzzing techniques. Such approaches struggle to detect semantically complex bugs that arise in real-world scenarios, particularly those involving nuanced interactions between application logic, physics engines, and user inputs. 

Moreover, existing efforts in game testing often prioritize test sequence generation or pattern-based test oracles~\cite{game_testing_2020}, approaches that are effective for detecting simple glitches but inadequate for uncovering subtle physics-related inconsistencies. \physicsbugs frequently manifest as subtle deviations from expected physical behaviors rather than immediate system failures, making them difficult to detect with traditional testing techniques. This results in many \physicsbugs remaining undetected until end-users encounter them, leading to negative reviews, decreased satisfaction, or even system failures in critical applications.

In this paper, we bridge this gap by exploring the effectiveness of detection techniques for physics failures in PE-based software systems. Specifically, we address the following research questions:

\textbf{RQ1:} What are the common manifestations of \physicsbugs in software systems that rely on PEs?

To answer this question, we conduct the first large-scale empirical study focused on identifying and categorizing \physicsbugs in PE-based software systems. We collect a diverse dataset of runtime behaviors from real-world software, including both buggy and non-buggy behaviors. This dataset comprises video evidence of \physicsbugs sourced from developer reports, user feedback, and community forums across multiple platforms. Through systematic analysis, we develop a comprehensive taxonomy of \physicsbug manifestations, identifying 17 distinct categories across 10 broader physics principles. Our findings reveal that gravity violations (40\%) and Newton's laws violations (28.2\%) dominate the distribution of \physicsbugs, followed by biomechanical failures (13.8\%) and collision detection failures (13.2\%).

\textbf{RQ2:} How effective are current techniques in detecting \physicsbugs from runtime behaviors? How do these techniques perform on multi-violation scenarios compared to single-violation cases?

For this research question, we evaluate the performance of a diverse set of state-of-the-art approaches in identifying \physicsbugs based on runtime behaviors. These approaches include deep learning-based video evaluation, prompt-based techniques, and large multimodal models (LMMs). We assess their capabilities across multiple dimensions, including violation detection and violation identification, on both the full \physicsbench dataset and the challenging \physicsbenchmulti subset containing 39 videos with concurrent violations. Our evaluation reveals that LMM-based methods generally outperform traditional computer vision approaches, with DEVIL-Gemini achieving the highest violation detection accuracy (69.2\%) and custom-designed Gemini prompts demonstrating exceptional violation identification capabilities (up to 89.5\% accuracy). Surprisingly, we find that \physicsbugs involving multiple simultaneous violations are often more readily detected than single-violation cases, with PhyGenEval showing a remarkable 9.3 percentage point accuracy improvement in multi-violation scenarios (76.3\% vs. 67.0\%). This counter-intuitive finding suggests that concurrent violations may create more pronounced deviations from expected behavior, though this benefit is exclusive to physics-aware approaches—general video anomaly detection methods show degraded performance on complex scenarios.

\textbf{RQ3:} How do developers perceive the status quo of \physicsbug detection during development and testing, and what are their perspectives on the effectiveness of current detection techniques?

To provide a practical perspective, we conduct a developer study to understand how practitioners currently handle \physicsbugs during software development and testing. Through a comprehensive questionnaire, we gather insights on the challenges developers face, their current approaches to identifying \physicsbugs, and their evaluation of the detection results produced by the tools examined in RQ2.

This paper makes the following contributions:
\begin{itemize}[leftmargin=*]
\item We present the first comprehensive empirical study of \physicsbugs in software systems that rely on PEs, based on curated datasets of runtime behavior videos (\physicsbench and \physicsbenchmulti). We develop a detailed taxonomy of \physicsbug manifestations across 17 distinct categories, revealing the prevalence and distribution of different physics violations in real-world software.

\item We conduct a thorough evaluation of current state-of-the-art methods in detecting \physicsbugs from runtime behaviors, demonstrating that while LMM-based approaches show promise (particularly with well-designed prompts), even the best methods still face significant challenges with subtle physics violations and context-dependent physics rules.

\item
We provide insights from developers on the practical challenges of handling \physicsbugs during development and testing, bridging the gap between research advancements and industry needs.
We provide insights on the practical challenges of handling \physicsbugs during testing,
bridging the gap between research advancements and industry needs.
Based on our findings, we offer recommendations for improving \physicsbug detection in PE-based software systems and outline promising directions for future research in this area.
\end{itemize}

%% file: sections/background.tex
\section{Preliminaries}
\label{sec:background}

\subsection{Problem Formulation}
\label{sec:problem_formulation}

We formalize the problem of detecting \physicsbugs as identifying deviations from expected physical behaviors through observable outputs without access to engine internals.

A PE-based software system $S$ consists of: (1) a \textbf{physics engine} $P$ simulating physical interactions, (2) \textbf{entities} $E = \{e_1, e_2, ..., e_n\}$ with physical properties, and (3) an \textbf{environment} $W$ defining world properties. During execution, the system produces observable states $O = (o_1, o_2, ..., o_T)$, typically as visual frames or sensor data.

A \textbf{\physicsbug} is an observable deviation that violates established physical principles. Let $B$ represent our taxonomy of \physicsbug categories. Each bug type $b \in B$ can be represented as a predicate over observation sequences: $b: (o_t, o_{t+1}, ..., o_{t+k}) \mapsto \{\text{true}, \text{false}\}$

Our detection problem seeks a function $D$ that evaluates whether observations $O$ contain evidence of any \physicsbug:$D(O) \mapsto \{\text{true}, \text{false}\} \times B'$, where $B' \subseteq B$ represents detected bug types. This task is challenging due to: (1) \textbf{semantic complexity} of bugs manifesting as behavioral patterns rather than crashes, (2) \textbf{observational limitations} from noisy and incomplete data, (3) \textbf{behavioral subtlety} requiring differentiation from intentional behaviors, and (4) \textbf{temporal dependencies} requiring sequence analysis rather than individual frames.

\subsection{Related Work: Existing Approaches That Have Potential to Solve This Problem}

\begin{table*}[t!]
\centering
\caption{Overview of methods and their characteristics}
\label{table:physics-bugs-empirical-approaches}
\resizebox{\linewidth}{!}{%
\begin{tabular}{lp{4cm}p{2cm}p{3cm}p{4cm}l}
\toprule
\textbf{Method Name} & \textbf{Title} & \textbf{Category} & \textbf{Base Model} & \textbf{Fine-tuned On} & \textbf{Model Size} \\ \midrule
\rowcolor{mygray} MIST~\cite{feng2021mist} & MIST: Multiple Instance Self-Training Framework for Video Anomaly Detection & Video anomaly detection & C3D/I3D feature extractors & ShanghaiTech/UCF-Crime & N/A \\
S3R~\cite{wu2022self} & Self-Supervised Sparse Representation for Video Anomaly Detection & Video anomaly detection & Feature encoder + dictionary learning & Self-supervised on anomaly datasets & N/A \\
\rowcolor{mygray}
Holmes-VAD~\cite{zhang2024holmes} & Holmes-VAD: Towards Unbiased and Explainable Video Anomaly Detection via Multi-modal LLM & Video anomaly detection & Llama3-Instruct-70B & VAD-Instruct50k & 70B parameters \\
VideoCon-Physics~\cite{bansal2024videophy} & VideoPhy: Evaluating Physical Commonsense for Video Generation & Video evaluation & VIDEOCON & VIDEOPHY dataset with semantic and physical labels & 7B parameters \\
\rowcolor{mygray}
VideoScore~\cite{he2024videoscore} & VideoScore: Building Automatic Metrics to Simulate Fine-grained Human Feedback for Video Generation & Video evaluation & Mantis-Idefics2 & VideoFeedback dataset & 8B parameters \\
DEVIL~\cite{liao2024evaluation} & Evaluation of Text-to-Video Generation Models: A Dynamics Perspective & Video evaluation & Gemini-1.5-Pro & None (prompt engineering only) & 175B parameters \\
\rowcolor{mygray}
PhyGenEval~\cite{meng2024world} & Towards World Simulator: Crafting Physical Commonsense-Based Benchmark for Video Generation & Video evaluation & GPT-4o and LLaVA & Physical commonsense evaluation dataset & Varies by component \\ 
\bottomrule
\end{tabular}
}
\vspace{-1em}
\end{table*}

To investigate the capability of existing techniques in detecting physics failures, we conduct rigorous systematic literature survey in top-tier (Core Ranking A and A*) computer science venues. 
At last, we identify three primary categories of approaches: deep learning-based video evaluation approaches, pure prompt engineering techniques leveraging large multimodal models (LMMs), and fine-tuned LMM-based approaches. 
We further adapt and investigate these techniques in Section~\ref{sec:rq2-method} and~\ref{sec:rq2-results}.

\subsubsection{Deep Learning Based Video Evaluation Approaches}
    (1) \textbf{MIST}~\cite{feng2021mist}: It refers to the Multiple Instance Self-Training Framework, which is a weakly supervised video anomaly detection (WS-VAD) framework that enhances video representations using a self-training approach. It focuses on generating reliable clip-level pseudo labels through sparse sampling and a self-guided attention-based feature encoder. MIST has demonstrated state-of-the-art performance on multiple benchmarks, achieving high accuracy in detecting anomalies with minimal supervision. We include MIST to investigate its applicability in detecting physics failures, which often manifest as subtle deviations in physical behavior within videos.
    (2) \textbf{S3R}~\cite{wu2022self}: It refers to Self-Supervised Sparse Representation for Video Anomaly Detection. It adopts a unified self-supervised framework that combines dictionary-based representations and self-supervised learning to localize anomalies in videos. It employs feature reconstruction and pseudo anomaly generation to effectively distinguish between normal and anomalous behaviors. The robustness of S3R across multiple video anomaly detection tasks makes it a strong candidate for identifying \physicsbugs in complex PE-based systems.

\subsubsection{Pure Prompt Engineering Approaches Based on General-purpose LMMs}
    (3) \textbf{Gemini}~\cite{gemini2023}. Gemini, a state-of-the-art general-purpose large multimodal model, has demonstrated impressive performance across a wide range of visual and textual tasks. By leveraging prompt engineering techniques, we explore Gemini's ability to identify \physicsbugs based on user-specified descriptions of physical phenomena and expected behavior. Gemini serves as a baseline to evaluate the capabilities of general-purpose LMMs in physics failure detection.
    (4) \textbf{DEVIL}~\cite{liao2024evaluation}. DEVIL focuses on assessing video content dynamics through a specialized evaluation protocol. It introduces metrics for temporal consistency, dynamics range, and dynamics controllability, making it a powerful tool for evaluating motion realism and physical correctness. While DEVIL primarily focuses on video dynamics assessment, it integrates with the Gemini multimodal model and employs specially designed prompts to evaluate naturalness, providing fine-grained assessment of whether videos adhere to physical laws in real-world contexts.

\subsubsection{Video Evaluation Approaches Based on Fine-tuned General-purpose Multi-modal Models}
    (5) \textbf{Holmes-VAD}~\cite{zhang2024holmes}. Holmes-VAD integrates multimodal instructions and precise temporal supervision to localize anomalies while providing human-like explanations for its predictions. It fine-tunes a multimodal large language model (LLM) on the VAD-Instruct50k dataset to generate interpretable content. Given its ability to detect and explain anomalies, Holmes-VAD is included to benchmark its performance in identifying complex physics failures.
(6) \textbf{VideoCon-Physics}~\cite{bansal2024videophy}. It is a video semantic and physical commonsense evaluation tool using the pre-trained VIDEOCON model, fine-tuned with the VIDEOPHY dataset. It assesses videos on two dimensions: Semantic Adherence (SA) and Physical Commonsense (PC), quantifying both consistency with textual descriptions and adherence to physical laws. We use this approach to test its ability to generalize to real-world physics failures in PE-based systems.
(7) \textbf{VideoScore}~\cite{he2024videoscore}.VideoScore is a video quality assessment tool using the pre-trained Mantis-Idefics2-8B visual-language model, fine-tuned on the VideoFeedback dataset. It evaluates video quality across five dimensions: Visual Quality, Temporal Consistency, Dynamic Degree, Text-to-Video Alignment, and Factual Consistency, enabling multi-dimensional analysis of both overall quality and adherence to semantic and physical laws. We use this method to assess its ability to detect physical failure phenomena in videos.

\subsubsection{Video Evaluation Approaches with Physics-Centric Fine-tuning}
    (8) \textbf{PhyGenEval}~\cite{meng2024towards}. PhyGenEval is a hierarchical evaluation framework that leverages advanced vision language models (VLMs) such as GPT-4o and LLaVA to assess whether videos conform to physical commonsense. The framework employs a three-layer structure that sequentially performs Key Physical Phenomena Detection, Physics Order Verification, and Overall Naturalness Evaluation, ultimately calculating an average score to comprehensively measure the video's overall physical commonsense performance. By applying PhyGenEval to runtime videos from PE-based systems, we evaluate its ability to detect deviations from physical norms that indicate \physicsbugs.

\noindent The combination of these approaches provides a comprehensive experimental foundation for evaluating the effectiveness of modern video anomaly detection techniques and LMM-based models in uncovering \physicsbugs. Our evaluation spans deep learning-based anomaly detection, prompt-driven reasoning, and fine-tuned multimodal learning, highlighting both the strengths and limitations of each approach in the context of PE-based systems.

%% file: sections/benchmark.tex
\section{\physicsbench Benchmark Construction}
\label{sec:benchmark_construction}

To systematically study and evaluate the detection of \physicsbugs in physics engine-based software systems, we constructed a comprehensive benchmark dataset consisting of runtime behaviors that exhibit both buggy and non-buggy physical phenomena. This section details the mechanical and objective process used to curate and preprocess the dataset, following established open coding procedures to minimize bias and ensure reproducibility.

\subsection{Data Collection}
We began by collecting instances of software exhibiting physics-related bugs from various gaming platforms, as these platforms provide a rich source of developer and user reports on physics anomalies. Specifically, we focused on platforms where large-scale games with complex physics systems are distributed, including \textit{Steam}, \textit{Epic}, \textit{Wegame}, and \textit{Xbox}. To identify relevant examples, we conducted keyword-based searches across these platforms, focusing on terms that are likely to correlate with \physicsbugs, such as \textit{game physics glitches}, \textit{physics simulation error bug}, and \textit{unity 3D physics glitches}.

After gathering initial game titles, we extended the search to community-driven platforms such as \textit{Reddit} and \textit{YouTube}, where users frequently share video clips highlighting game bugs. To broaden the coverage of \physicsbug instances, we utilized YouTube's recommendation algorithm to discover related videos, starting from the ones initially found via keyword searches. Additionally, we cross-referenced our findings with examples from the \textit{GlitchBench} dataset~\cite{taesiri2024glitchbench}, selecting relevant cases where the original sources were still available.

Throughout this process, we carefully annotated the corresponding video timestamps and links using manual labeling, ensuring precise documentation of each example. The collected video data was downloaded using \textit{yt-dlp}~\cite{ytdlp} and processed using \textit{moviepy}~\cite{moviepy} to isolate relevant segments.%

\subsection{Data Preprocessing}
After collecting the raw video data, we performed a series of preprocessing steps to ensure consistency and relevance across the dataset. Each video was manually inspected and trimmed to remove unrelated frames, such as initial and ending segments that did not involve any gameplay or physics simulations. Additionally, audio tracks were removed to focus solely on visual representations of physics behaviors.
Next, the dataset underwent a rigorous filtering process. Following exclusion criteria were applied: low resolution, extraneous content, perspective shifts, and obstructed view.

Following this preprocessing, we arrived at a final dataset, \physicsbench, comprising 1,000 video clips. These include 500 instances where the game adhered to real-world physical laws (\textit{non-buggy videos}) and 500 instances of physics-related bugs (\textit{buggy videos}). The non-buggy videos were all sourced from YouTube, while the buggy videos were distributed across multiple platforms: 429 from YouTube, 54 from Reddit, and 17 from the GlitchBench dataset.
Among all data in \physicsbench, there are 39 videos that contains more than one \physicsbug at the same time which may make the detection process more difficult. These data form a subset of \physicsbench, we call them \physicsbenchmulti.

%% file: sections/taxonomy.tex
\section{RQ1: Manifestation Taxonomy}
We first design a comprehensive taxonomy on the \physicsbugs to systematically categorize the observed manifestations based on violations of core principles of physics. 
Then, we further analyze the different manifestation categories of \physicsbug.

\subsection{Taxonomy Design}
Once the video dataset is curated, we conduct a detailed classification of the physical phenomena depicted in each video. Specifically, we categorized each instance of a \physicsbug according to the violated physical laws, such as Newtonian mechanics, rigid body collisions, or fluid dynamics. 
To ensure objectivity and consistency, we adapt the widely-adopted open coding procedure~\cite{book:open-coding-16} and employ a collaborative classification method involving both human experts and AI assistance. Initially, a set of physics rule categories was defined by human annotators, and these initial annotations were iteratively refined using a large multimodal model (LMM) such as GPT-4o. The model's reflective capability allowed it to critique and improve its own classifications based on feedback loops, ensuring a thorough exploration of edge cases. To further validate our classification schema, we consulted two experts in physics, who provided an external review of the rationality and completeness of the final categories.

Through this process, the 500 buggy videos were divided into \numberOfPhysicsBugCategories distinct categories of physics violations as shown in Table~\ref{tab:category_ratio}. 
This classification forms the foundation for subsequent evaluations, enabling a structured analysis of the types and frequency of \physicsbugs present in the dataset.

\setlength{\subfigbottomskip}{-0.8pt}
\begin{figure*}[t!]
\vspace{-1em}
\centering

    \subfigure[Weightlessness]{
        \label{fig:Weightlessness}
        \includegraphics[width=0.485\textwidth]{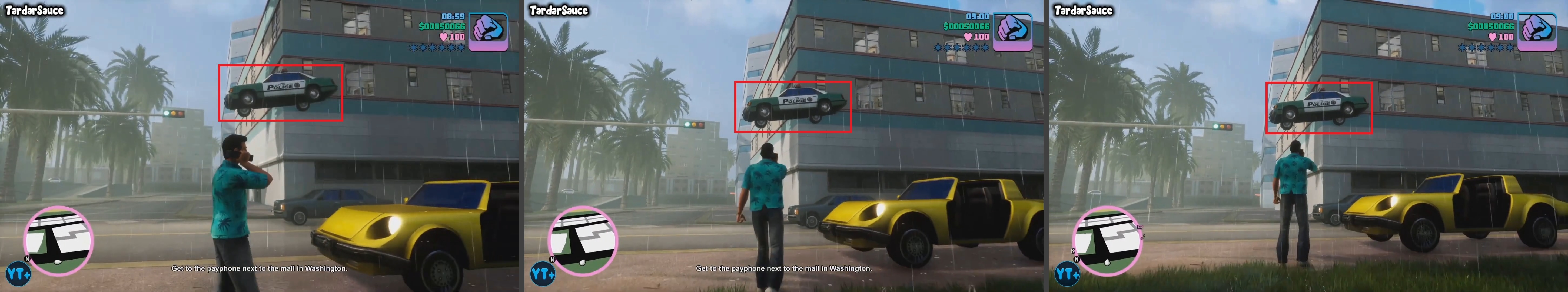}
    }
    \hspace{-1em}
    \subfigure[Anti-Gravity]{
        \label{fig:Anti-Gravity}
        \includegraphics[width=0.485\textwidth]{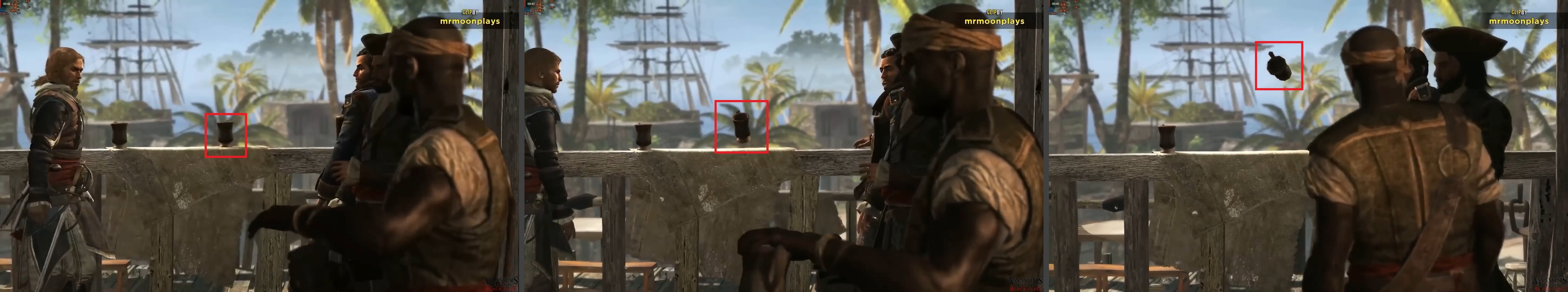}
    }

    \subfigure[Delayed Gravity Effect]{
        \label{fig:Delayed Gravity Effect}
        \includegraphics[width=0.485\textwidth]{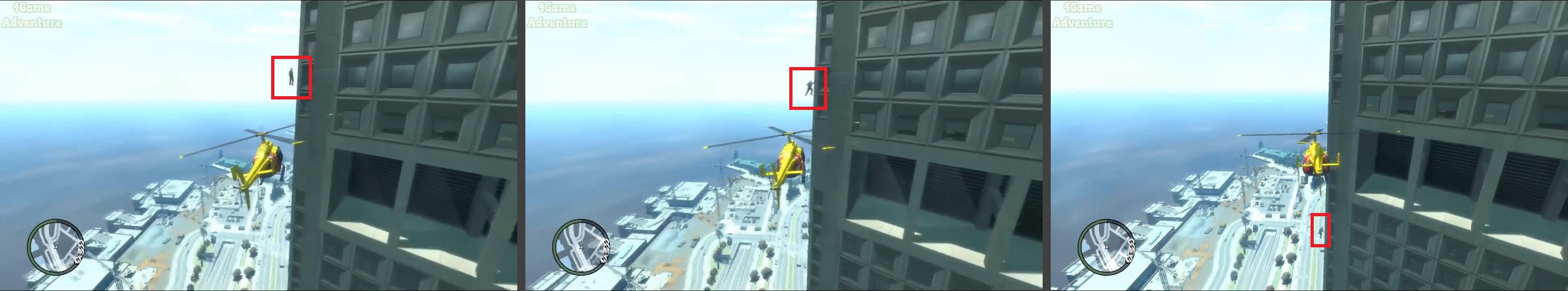}
    }
    \hspace{-1em}
    \subfigure[Clipping-Through (Phantom Overlap)]{
        \label{fig:Clipping-Through (Phantom Overlap}
        \includegraphics[width=0.485\textwidth]{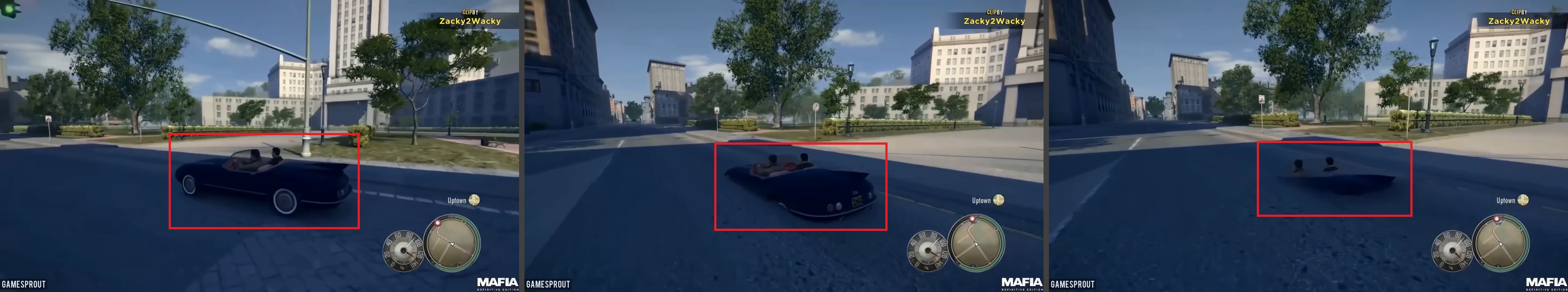}
    }

    \subfigure[Spontaneous Rapid Spinning]{
        \label{fig:Spontaneous Rapid Spinning}
        \includegraphics[width=0.485\textwidth]{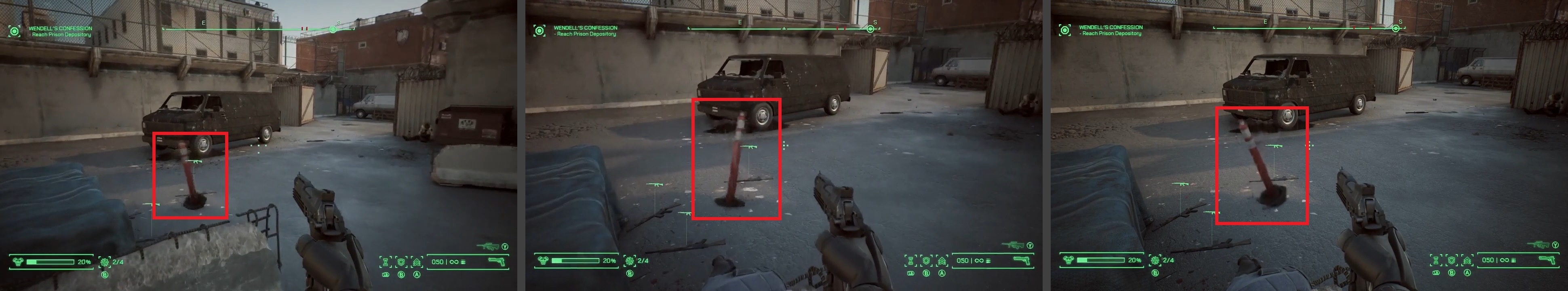}
    }
    \hspace{-1em}
    \subfigure[Sudden Appearance or Disappearance]{
        \label{fig:Sudden Appearance or Disappearance}
        \includegraphics[width=0.485\textwidth]{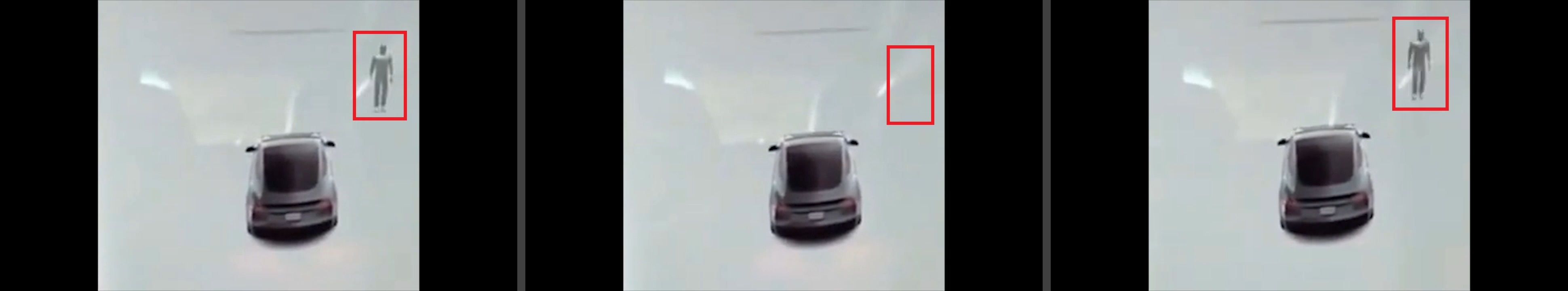}
    }

    \subfigure[Spontaneous Motion]{
        \label{fig:Spontaneous Motion}
        \includegraphics[width=0.485\textwidth]{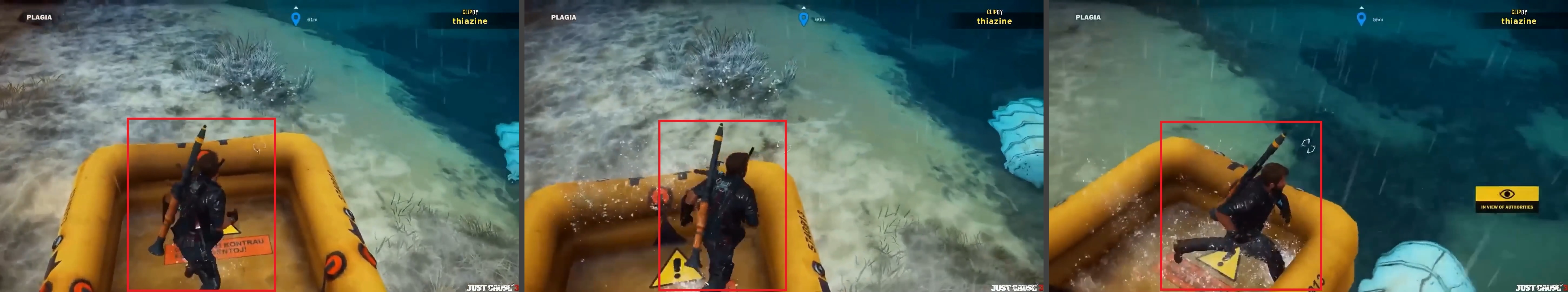}
    }
    \hspace{-1em}
    \subfigure[Uncaused Directional Change]{
        \label{fig:Uncaused Directional Change}
        \includegraphics[width=0.485\textwidth]{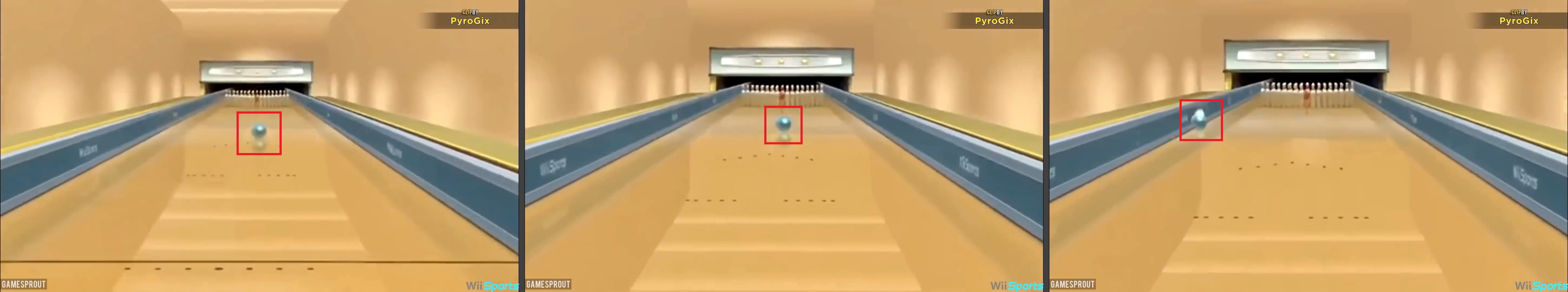}
    }

    \subfigure[Instantaneous State Transition]{
        \label{fig:Instantaneous State Transition}
        \includegraphics[width=0.485\textwidth]{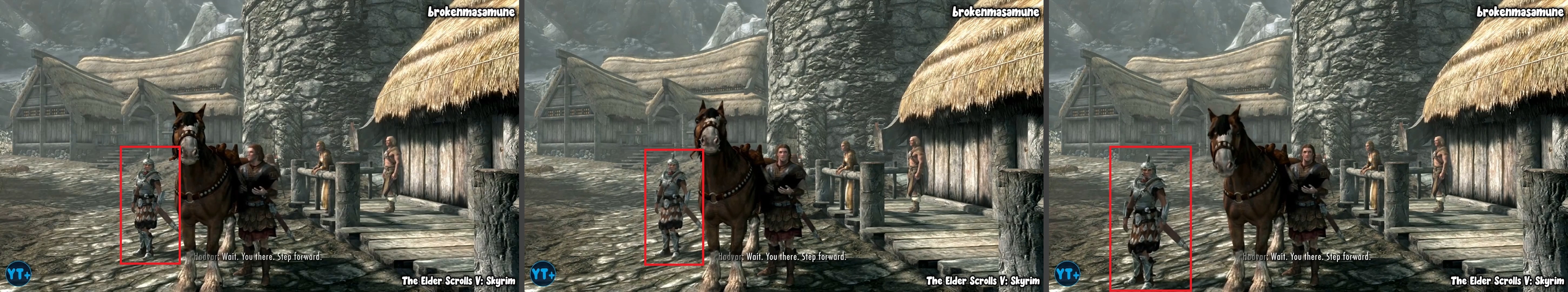}
    }
    \hspace{-1em}
    \subfigure[Incorrect Force-Acceleration Relation]{
        \label{fig:Incorrect Force-Acceleration Relation}
        \includegraphics[width=0.485\textwidth]{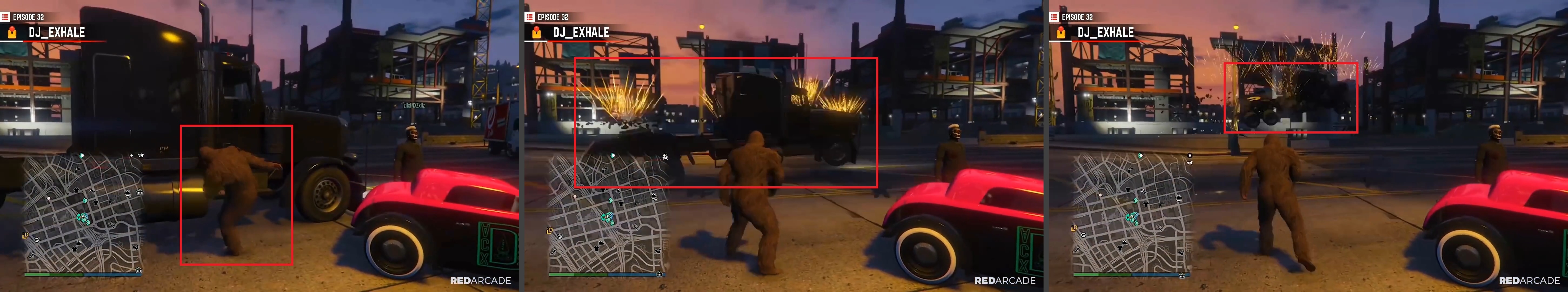}
    }

    \subfigure[Biomechanical Failures]{
        \label{fig:Biomechanical Failures}
        \includegraphics[width=0.485\textwidth]{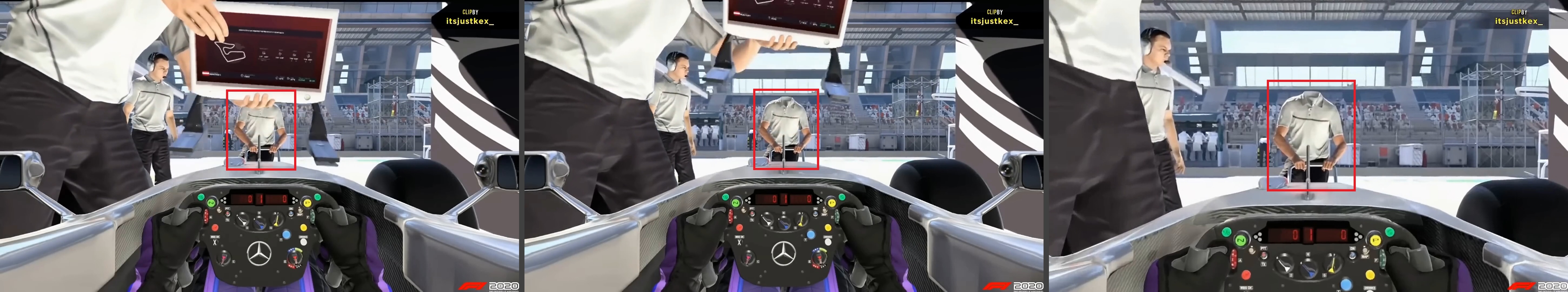}
    }
    \hspace{-1em}
    \subfigure[Buoyancy Violations]{
        \label{fig:Buoyancy Violations}
        \includegraphics[width=0.485\textwidth]{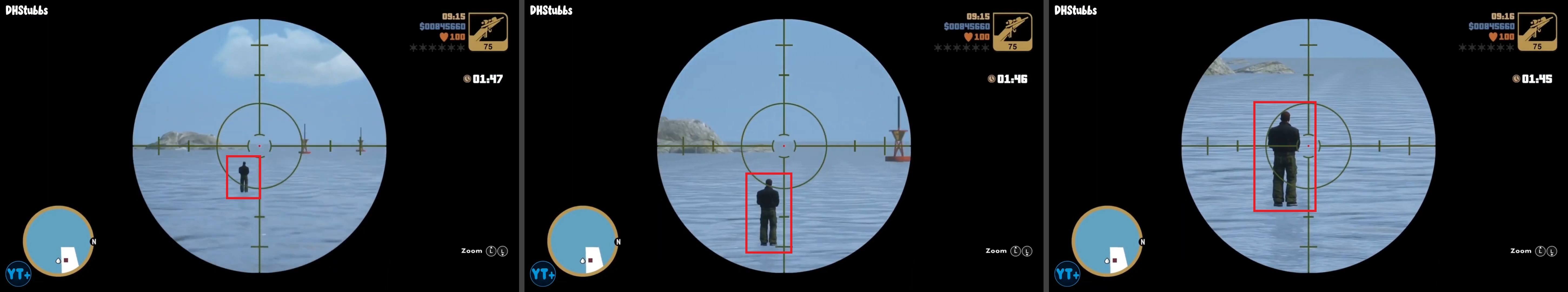}
    }

    \subfigure[Fluid Dynamics Failures]{
        \label{fig:Fluid Dynamics Failures}
        \includegraphics[width=0.485\textwidth]{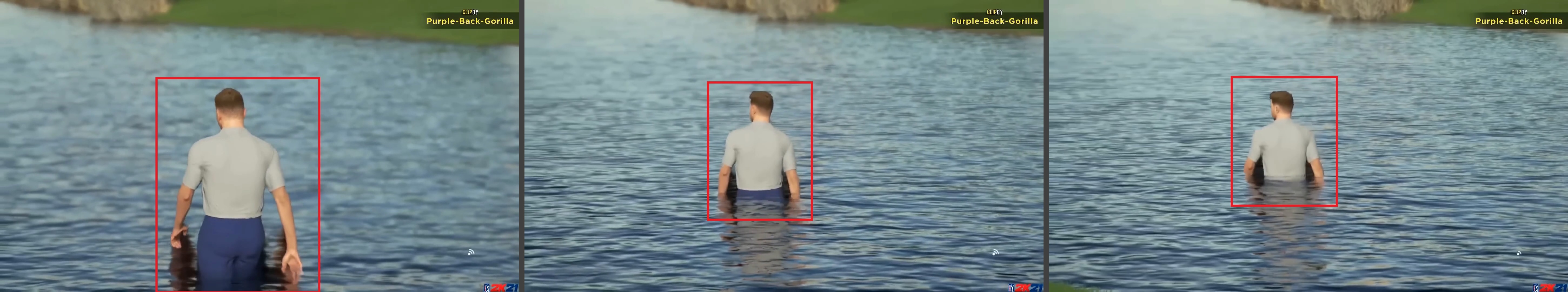}
    }
    \hspace{-1em}
    \subfigure[Aerodynamics Violations]{
        \label{fig:Aerodynamics Violations}
        \includegraphics[width=0.485\textwidth]{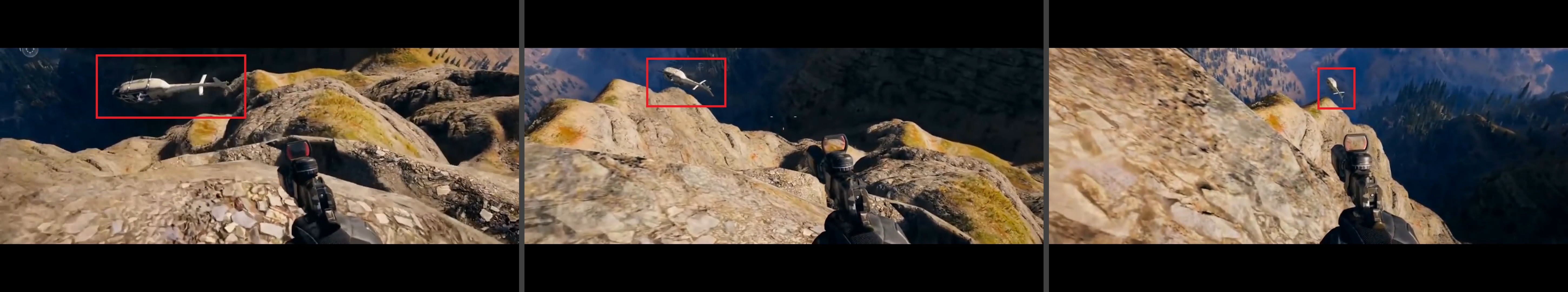}
    }

    \subfigure[Thermodynamic Anomalies]{
        \label{fig:Thermodynamic Anomalies}
        \includegraphics[width=0.485\textwidth]{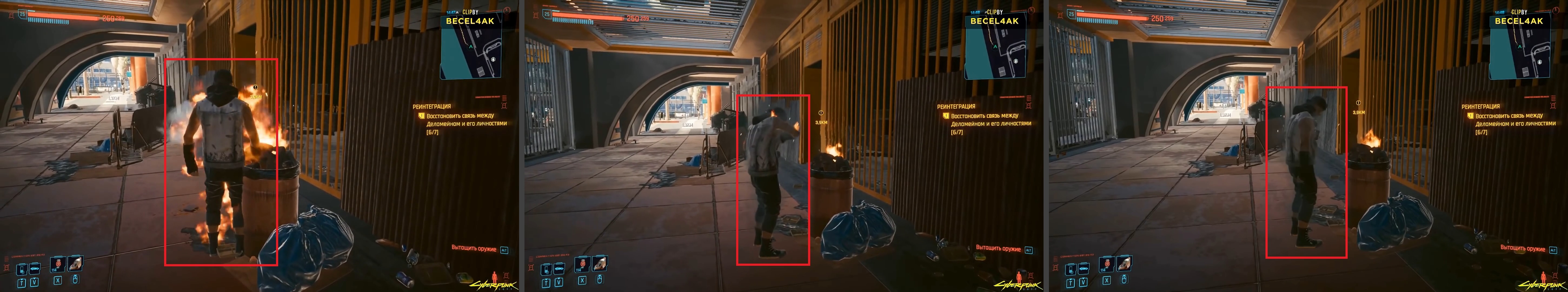}
    }
    \hspace{-1em}
    \subfigure[Optical Physics Violations]{
        \label{fig:Optical Physics Violations}
        \includegraphics[width=0.485\textwidth]{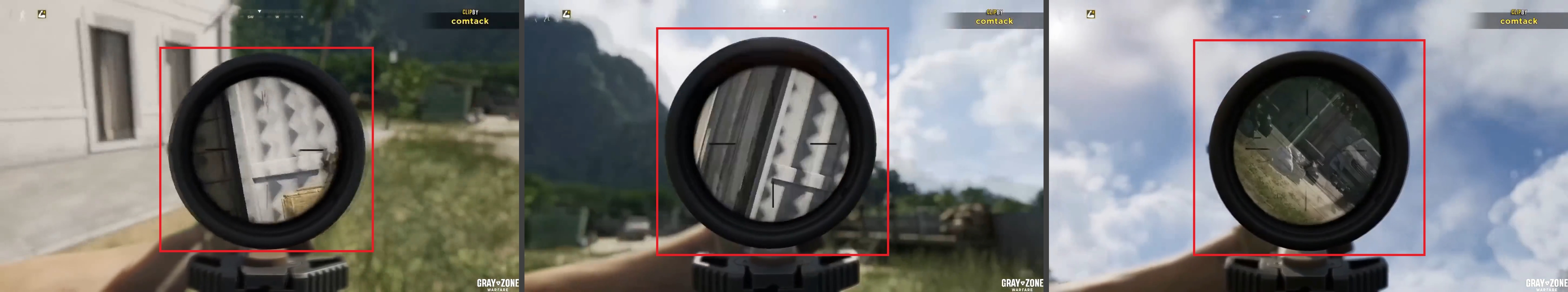}
    }

    \subfigure[Structural and Material Mechanics Failures]{
        \label{fig:Structural and Material Mechanics Failures}
        \includegraphics[width=0.485\textwidth]{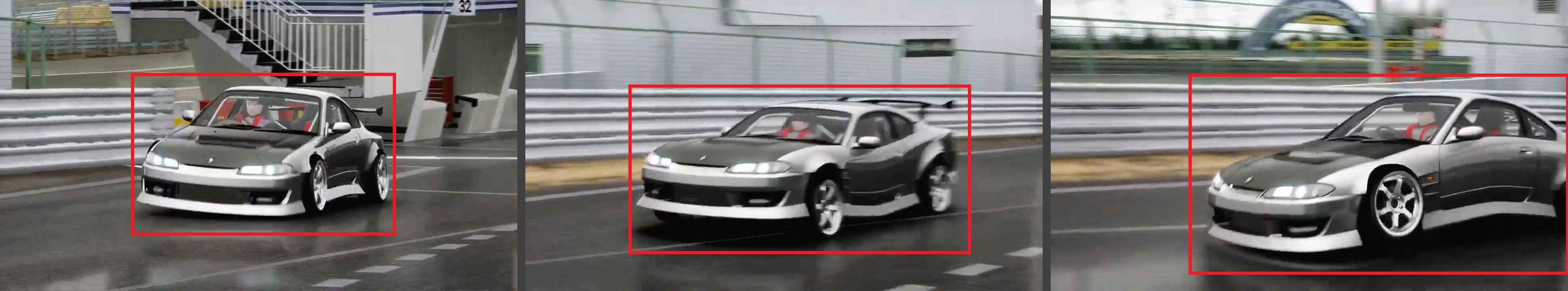}
    }

\caption{Diversified manifestation categories of \physicsbugs. Examples are demonstrated with three frames in time order to reflect the dynamic failures. The portions of physics violations are highlighted with {\color{red} red boxes}.}
\label{fig:double_column_example}
\vspace{-1em}
\end{figure*}

%
%

\begin{table}[t!]
\centering
\caption{Manifestation Category Distribution of \physicsbench\todoaftersub{The Collision category??}}
\label{tab:category_ratio}
\resizebox{0.8\columnwidth}{!}{%
\begin{tabular}{lr}
\toprule
\textbf{Manifestation Category} & \textbf{Ratio (\%)} \\
\midrule
\textbf{Gravity Violations} & \textbf{40.0} \\
\midrule
\rowcolor{mygray}
Weightlessness & 22.0 \\
Anti-Gravity & 16.0 \\
\rowcolor{mygray}
Delayed Gravity Effect & 2.0 \\
\midrule
\textbf{Collision Detection Failures} & \textbf{13.2} \\
\midrule
Clipping-Through (Phantom Overlap) & 13.2 \\
\midrule
\textbf{Newton’s Laws Violations} & \textbf{28.2} \\
\midrule
\rowcolor{mygray}
Spontaneous Rapid Spinning & 5.0 \\
Sudden Appearance or Disappearance & 4.8 \\
\rowcolor{mygray}
Spontaneous Motion & 4.8 \\
Uncaused Directional Change & 1.4 \\
\rowcolor{mygray}
Instantaneous State Transition & 1.2 \\
Incorrect Force-Acceleration Relation & 11.0 \\
\midrule
\textbf{Biomechanical Failures} & \textbf{13.8} \\
\midrule
\textbf{Buoyancy Violations} & \textbf{1.2} \\
\midrule
\textbf{Fluid Dynamics Failures} & \textbf{1.4} \\
\midrule
\textbf{Aerodynamics Violations} & \textbf{0.8} \\
\midrule
\textbf{Thermodynamic Anomalies} & \textbf{0.4} \\
\midrule
\textbf{Optical Physics Violations} & \textbf{0.4} \\
\midrule
\textbf{Structural and Material Mechanics Failures} & \textbf{0.6} \\
\bottomrule
\end{tabular}
}
\end{table}

\subsection{Details of Manifestation Categories}
Here, we provide a detailed description of each manifestation category, along with examples to enhance understanding.
\subsubsection{Gravity Violations} 
Gravity is a fundamental force that governs the behavior of objects with mass. Several types of physics failures occur when simulated objects violate the principles of gravity as follows:
\begin{itemize}[leftmargin=*]
    \item \textbf{Weightlessness:} Objects remain stationary, move uniformly, or drift erratically in midair without any external force, violating the principle of gravity. For example, Figure~\ref{fig:Weightlessness} shows a police car hovering above a city street without visible support.
    \item \textbf{Anti-Gravity:} Objects accelerate upward without any external force acting against gravity, contradicting the expected direction of gravitational pull.

    \item \textbf{Delayed Gravity Effect:} Gravity is delayed or temporarily suspended, causing objects to remain suspended in the air for a period before falling, which defies the principle of instantaneous gravitational influence. %
\end{itemize}

\subsubsection{Collision Detection Failures} 
Collision detection, a core functionality in PEs, ensures that objects cannot occupy the same space simultaneously. Failures in collision detection violate the laws of object exclusion and impenetrability:
\begin{itemize}[leftmargin=*]
    \item \textbf{Clipping-Through (Phantom Overlap):} Objects fail to register collisions upon contact, resulting in mutual penetration or embedding. This behavior violates the law of impenetrability.
\end{itemize}

\subsubsection{Newton's Laws Violations} Newton's laws of motion are foundational to mechanics, and several physics failures can be categorized based on violations of these laws:
\begin{itemize}[leftmargin=*]
    \item \textbf{Spontaneous Rapid Spinning:} Objects begin to spin rapidly without the presence of external torque or energy input, violating the conservation of angular momentum and mechanical energy. For example, a cone-shaped barrier that spontaneously begins spinning at high speed without any external force.

    \item \textbf{Sudden Appearance or Disappearance:} Objects materialize or vanish without following the law of conservation of mass and matter.

    \item \textbf{Spontaneous Motion:} Objects accelerate or decelerate without any external force, violating Newton’s First Law of Motion. For example, Figure~\ref{fig:Spontaneous Motion}, a person running inside a stationary kayak causes the kayak to move forward, incorrectly treating internal forces as external.

    \item \textbf{Uncaused Directional Change:} Objects change direction abruptly without external force, violating the principle of inertia.

    \item \textbf{Instantaneous State Transition:} The movement or position of an object undergoes a sudden, discontinuous jump, violating the continuity of space and time. For example, a soldier teleporting instantaneously to a distant location.

    \item  \textbf{Incorrect Force-Acceleration Relation:} The net force applied to an object does not match the product of its mass and acceleration, violating Newton’s Second Law.
\end{itemize}

\subsubsection{Biomechanical Failures} Biomechanical failures occur when the motion, posture, or anatomy of biological entities violate fundamental biological principles. For example, Figure~\ref{fig:Biomechanical Failures} shows a pit crew member in a racing game missing their head, violating basic anatomical rules.

\subsubsection{Buoyancy Violations} 
Buoyancy, governed by an object’s density and fluid properties, is violated when objects do not float or sink in accordance with these factors.

\subsubsection{Fluid Dynamics Failures} 
Fluid dynamics involves the interaction of liquids with objects and other fluids. Physics failures in this area occur when liquid motion or interaction defies physical laws governing fluid behavior. For example, a person walks into a lake, but the water's surface shows no ripples or disturbance.

\subsubsection{Aerodynamics Violations} 
Aerodynamics governs the motion of objects in air, including drag, lift, and fluid flow. Failures in aerodynamics are observed when objects behave in ways that contradict these principles.

\subsubsection{Thermodynamic Anomalies} 
Thermodynamics relates to heat transfer, energy conversion, and changes in entropy. Failures in thermodynamics occur when systems violate these laws. For example, a person on fire suddenly extinguishing without any external cause, violating conservation of energy and heat transfer principles.

\subsubsection{Optical Physics Violations} 
Optical physics governs the behavior of light, including reflection, refraction, and linear propagation. Failures in this domain include violations of these principles, such as incorrect light propagation. Figure~\ref{fig:Optical Physics Violations} shows a telescope aimed at the sky displaying an image inside that does not match the external view.

\subsubsection{Structural and Material Mechanics Failures} 
Structural and material mechanics govern how materials behave under stress and deformation, guided by principles such as Hooke’s Law and Young’s Modulus. Failures in this area include unrealistic material deformations under force. For example, a car’s body continuously twisting and deforming while driving, violating principles of structural integrity and elasticity.

This taxonomy provides a detailed classification of physics failures observed in simulations. Understanding these failures is crucial for improving the robustness of physics simulation engines and ensuring the fidelity of simulations in critical applications like autonomous systems, gaming, and virtual reality. 
By systematically categorizing these failures, our work offers a foundation for future research on the detection and mitigation of \physicsbugs in PEs.

%% file: sections/experiment.tex
\input{sections/figs}

\section{Research Methodology for RQ2}%
\label{sec:rq2-method}

\subsection{Adaptation of Existing Approaches for Physics Failure Detection}

We adapt a variety of state-of-the-art video evaluation and anomaly detection approaches to identify and classify \physicsbugs in runtime behaviors of PE-based systems. Our experimental goals are twofold: (1) \textbf{Violation Detection (VD)}, which is to evaluate whether each method can correctly determine if a video violates physical rules. (2) \textbf{Violation Identification (VI)}, which is to assess whether each method can identify which specific physical rule is violated.

The metrics generated by these methods primarily fall into two categories: (1) \textbf{Physics Correctness (PC)}: This metric directly evaluates adherence to physical rules. (2) \textbf{Semantic Alignment (SA)}: We design video description texts that include physical rules and evaluate how well the text aligns with the video to determine whether the rules mentioned in the text are followed.

Here, we describe specific configurations for each approach:
\begin{itemize}[leftmargin=*]
    \item \textbf{VideoPhy}: We conduct a two-phase evaluation process: For physics correctness (PC), we perform a single evaluation pass using only video input to detect physical law violations (VD). For semantic alignment (SA), three sequential rounds are executed with progressively refined prompts that include: (1) integrating specific rules into video content descriptions (e.g., ``The helicopter hovers near the mountain top, staying aloft due to principles of aerodynamics...''), (2) providing only specific rules (e.g., ``This video follows the physics law of Aerodynamics''), and (3) explicitly referencing physical laws generally (e.g., ``This video follows all physics laws''). Results from rounds 1 and 2 are used to identify which specific physical law is violated (VI), while round 3 results determine whether any physical laws are violated (VD).
    \item \textbf{VideoScore}: We adapt VideoScore using the same three-round prompt structure as VideoPhy. Of the five dimensions VideoScore evaluates (Visual Quality, Temporal Consistency, Dynamic Degree, Text-to-Video Alignment, and Factual Consistency), we focus on Text-to-Video Alignment as our SA metric and Factual Consistency as our PC metric. For PC evaluation, we use the third round results to assess whether videos violate physical laws (VD). For SA evaluation, we use first and second round results to identify specific violated rules (VI) and third round results to determine overall violation status (VD).
    \item \textbf{DEVIL}: For evaluating video physics correctness using DEVIL, we leverage only its Naturalness metric, originally computed via the Gemini large multimodal model (LMM). To further validate generalizability, we perform a evaluation round by replacing Gemini with Holmes-VAD. Thus, we conducted two independent rounds of experiments using Gemini and Holmes. Each round uses identical prompt configurations, focusing exclusively on determining whether videos violate physical laws (VD) based solely on the Naturalness score generated by the respective LMMs.
    \item \textbf{PhyGenEval}: We apply PhyGenEval using the same three rounds of prompts identical to VideoPhy and VideoScore. PhyGenEval's hierarchical framework ultimately generates a comprehensive score measuring overall physical commonsense performance. Results from rounds 1 and 2 are used to identify which specific physical law is violated (VI), while round 3 results determine whether any physical laws are violated (VD).
    \item \textbf{MIST-VAD}: MIST-VAD evaluates physics correctness exclusively. We configured experiments across two datasets (ShanghaiTech and UCF-Crime) and two feature encoders (C3D and I3D), generating four result sets: \textit{mistvad\_sht\_c3d}, \textit{mistvad\_sht\_i3d}, \textit{mistvad\_ucf\_c3d}, and \textit{mistvad\_ucf\_i3d}. All four results are used to evaluate whether videos violate physical laws (VD).
    \item \textbf{S3R}: S3R is adapted for physics correctness evaluation only, with a single round of experiments producing results to determine whether videos violate physical laws (VD).
    \item \textbf{Gemini}: We evaluate Gemini using two distinct approaches: For \textbf{Standard Prompt Structure}, we use the same three-round prompt structure as VideoPhy for comparative analysis. For \textbf{Gemini-Specific Prompts}, three custom rounds are conducted: (1) freeform reasoning where the model generates its own rules and determines violations, (2) rule-based classification using our standard rule categories, and (3) rule-based classification with examples following few-shot principles. The results of all three rounds are used to identify which specific physical law is violated (VI).
    \item \textbf{Holmes-VAD}: Holmes-VAD is a general multimodal model, but it cannot follow the format requirements of the instructions to output accurate scores. We therefore apply the same three-round custom prompt approach used with Gemini's specific prompts, focusing on qualitative evaluation of physics violations (VI).
\end{itemize}

\subsection{Experimental Setup}

We design our experimental setup to benchmark the performance of selected approaches in detecting \physicsbugs within PE-based systems, conducted in the following controlled environment:

\subsubsection{Evaluation Metrics}

For comprehensive evaluation, we employ both classification metrics and correlation-based metrics. The classification metrics include Accuracy, Precision, Recall, F1 Score, and AUC-ROC.

\begin{equation}
\text{AUC-ROC} = \int_{0}^{1} \text{TPR}(t) \times \text{FPR}'(t) \, dt
\end{equation}

To assess alignment between model predictions and ground truth values, we calculate correlation coefficients:

\begin{equation}
r = \frac{\sum_{i=1}^{n} (x_i - \bar{x})(y_i - \bar{y})}{\sqrt{\sum_{i=1}^{n} (x_i - \bar{x})^2 \sum_{i=1}^{n} (y_i - \bar{y})^2}}
\quad \quad
\rho = 1 - \frac{6 \sum d_i^2}{n(n^2 - 1)}
\end{equation}

where $r$ is the Pearson correlation coefficient measuring linear relationships, and $\rho$ is the Spearman rank correlation coefficient assessing monotonic relationships. For all metrics, higher values indicate better performance, except negative correlations which suggest inverse relationships between predictions and ground truth.

\subsubsection{Adaptation Perspectives} To evaluate the effectiveness of each approach on \physicsbug detection, we employ two primary perspectives:
\begin{itemize}[leftmargin=*]
        \item \textbf{Physics Correctness (PC):} Whether the video content adheres to the laws of physics, as determined by task-specific outputs (e.g., flags or violation categories).
        \item \textbf{Semantic Alignment (SA):} A measure of how well the generated description aligns with the video content. This is computed across multiple prompt iterations when applicable.
    \end{itemize}

\subsubsection{Experimental Data} We curate a dataset consisting of buggy and non-buggy video instances sourced from platforms such as Steam, Reddit, and YouTube. These videos capture diverse physics failure scenarios, including gravity violations, collision detection failures, and Newton's law violations.

\subsubsection{Evaluation Procedure} Each approach is executed according to the adaptation strategy described in Section 3.2. Specifically:
    \begin{itemize}[leftmargin=*]
        \item \textbf{Multi-Prompt Strategies:} For approaches such as \textit{VideoPhy}, \textit{VideoScore}, and \textit{PhyGenEval}, we execute multiple rounds of evaluation with progressively refined prompts to capture both physics correctness and semantic alignment.
        \item \textbf{Single-Prompt Execution:} Methods like \textit{DEVIL} and \textit{S3R} operate using a single prompt, focusing on their intrinsic evaluation capabilities.
        \item \textbf{Variant Outputs:} For approaches producing multiple output datasets (e.g., \textit{MIST-VAD}), we collect results across all configurations to ensure comprehensive evaluation.
    \end{itemize}

    \subsubsection{Experimental Tools} All experiments are performed on a high-performance computing cluster equipped with NVIDIA GPUs and TensorFlow/PyTorch backends for deep learning models. The prompt-based evaluations are conducted using LMM APIs such as Gemini and Holmes-VAD.

\noindent By leveraging diverse approaches and a carefully designed evaluation strategy, our experimental setup aims to provide a rigorous benchmark for identifying and characterizing \physicsbugs in PE-based systems.

\input{sections/results_analysis}

%% file: sections/figs.tex
\begin{figure*}[t!]
\centering
  \begin{minipage}[b]{0.245\linewidth}
    \centering
    \includegraphics[width=\linewidth]{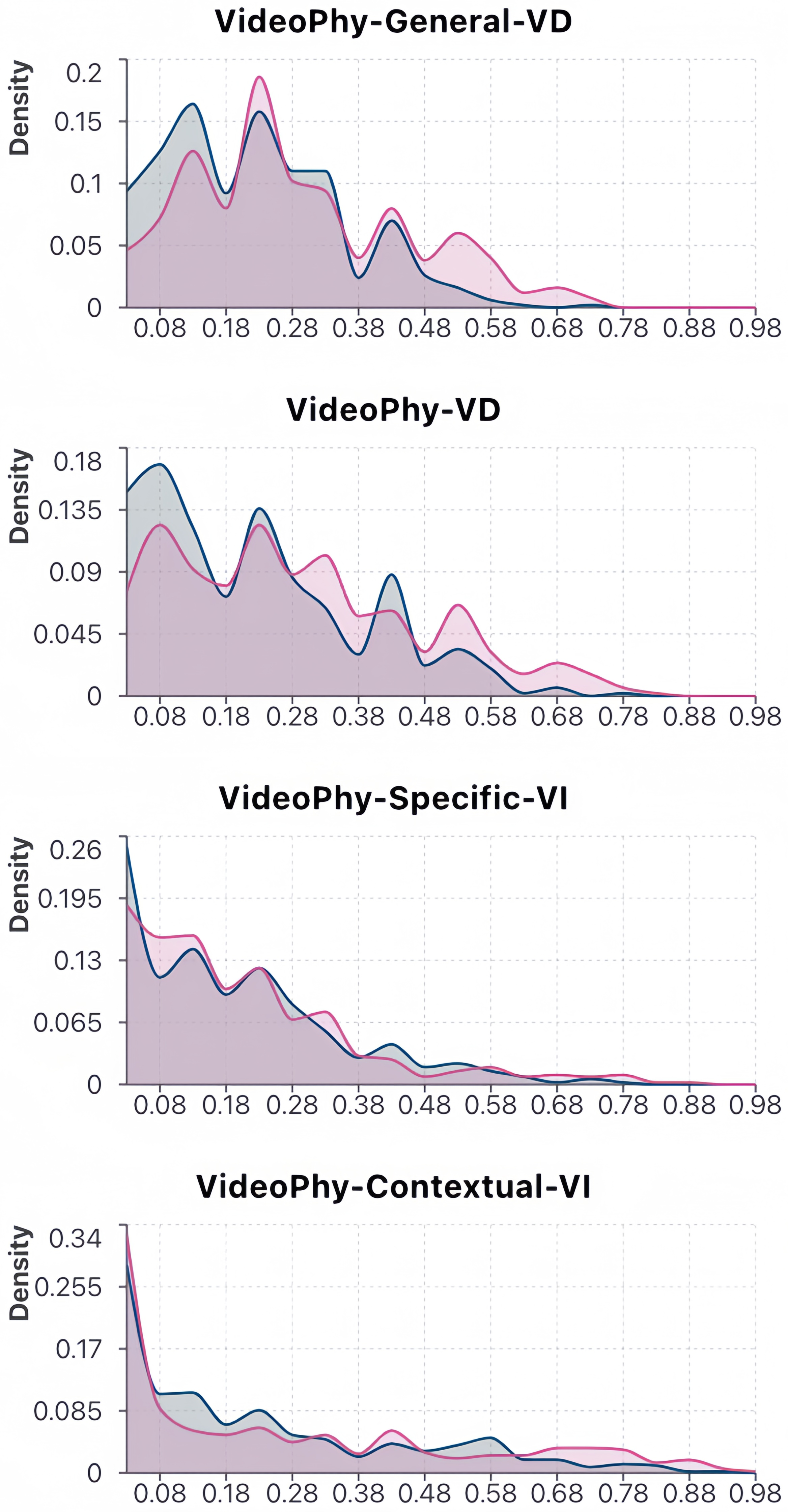}
    \caption{VideoPhy}
    \label{fig2}
  \end{minipage}%
  \hfill%
  \begin{minipage}[b]{0.245\linewidth}
    \centering
    \includegraphics[width=\linewidth]{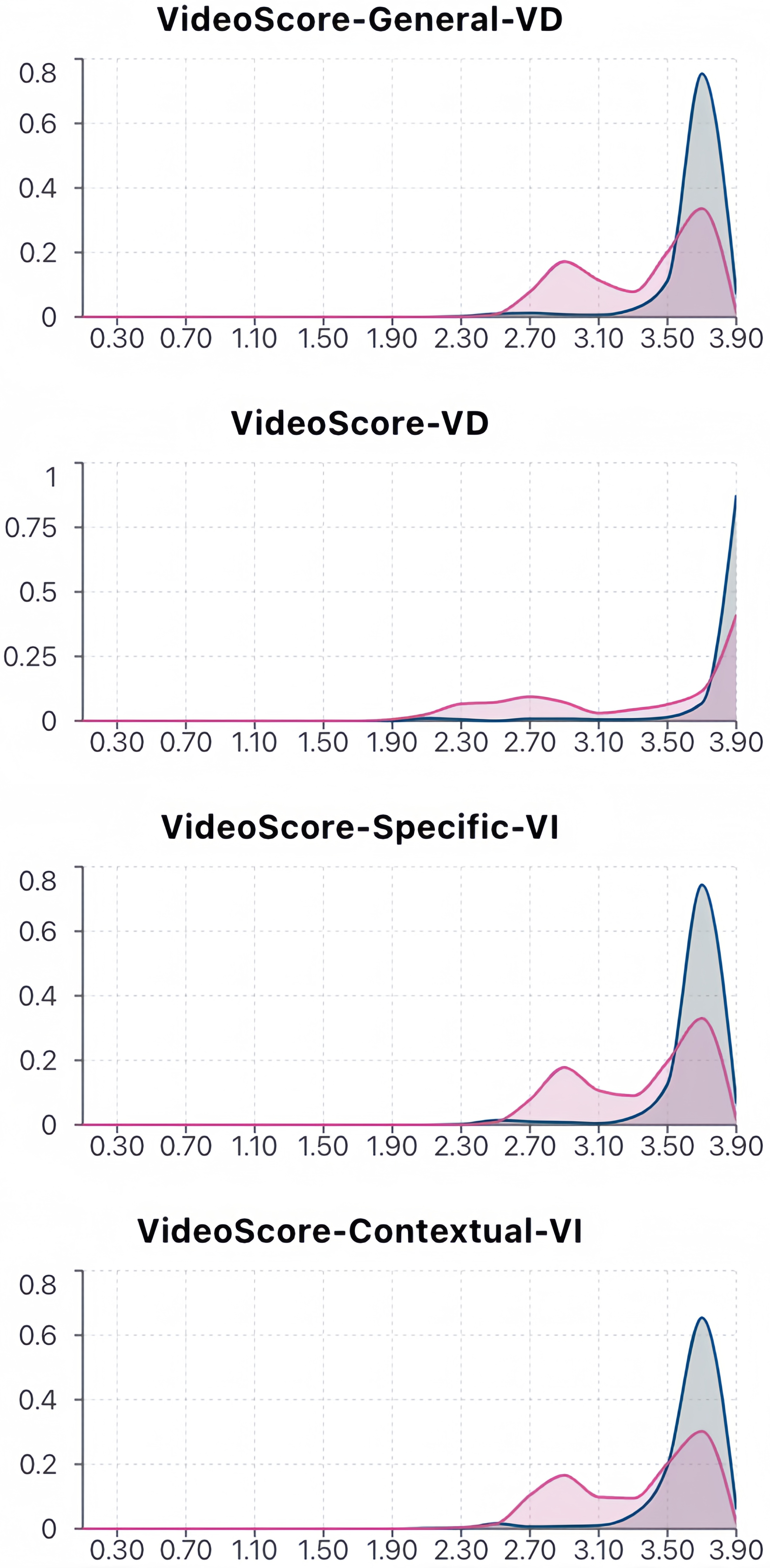}
    \caption{VideoScore}
    \label{fig3}
  \end{minipage}%
  \hfill%
  \begin{minipage}[b]{0.245\linewidth}
    \centering
    \includegraphics[width=\linewidth]{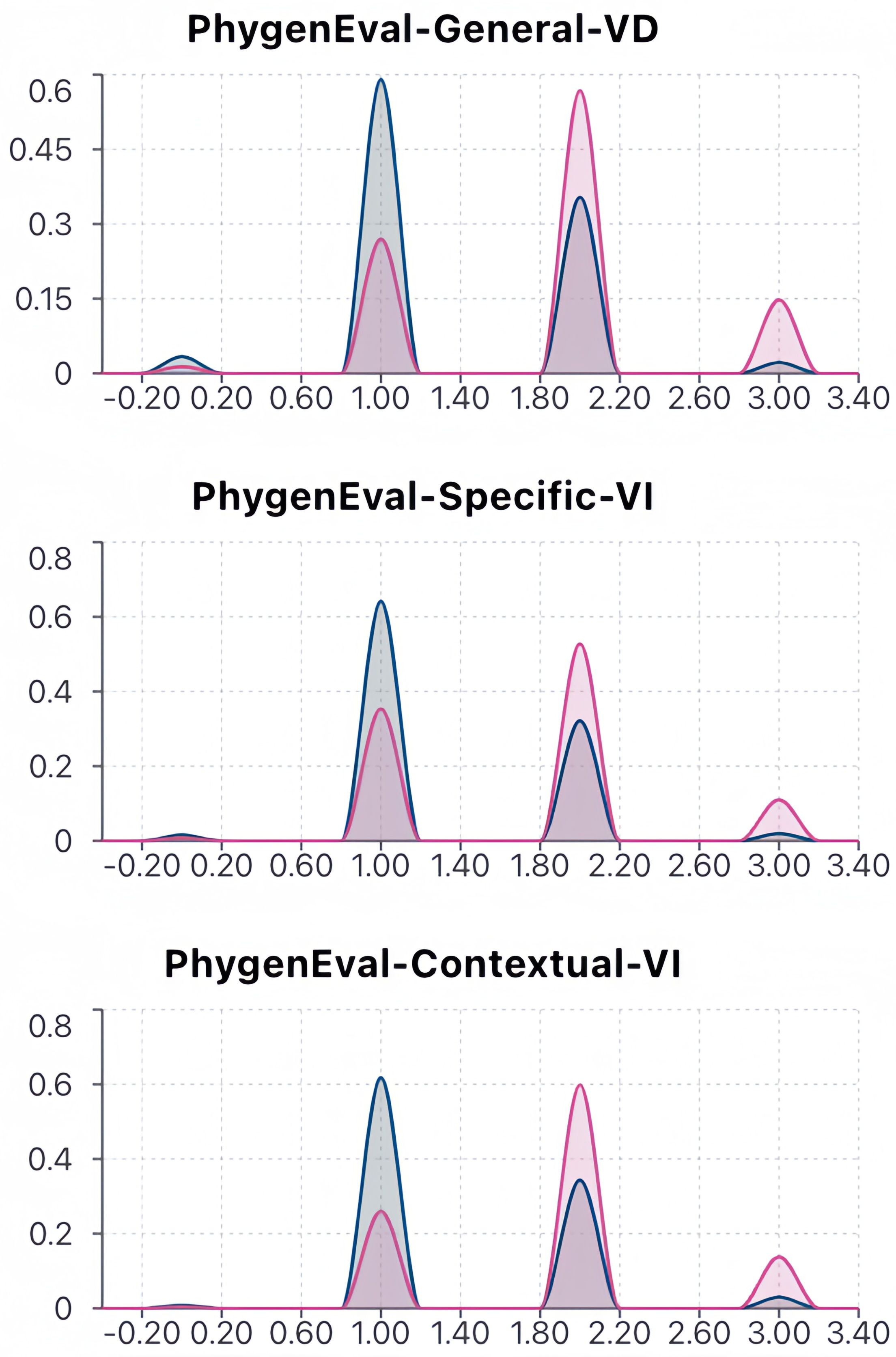}
    \caption{PhygenEval}
    \label{fig4}
  \end{minipage}%
  \hfill%
  \begin{minipage}[b]{0.245\linewidth}
    \begin{minipage}[b]{\linewidth}
      \centering
      \includegraphics[width=\linewidth]{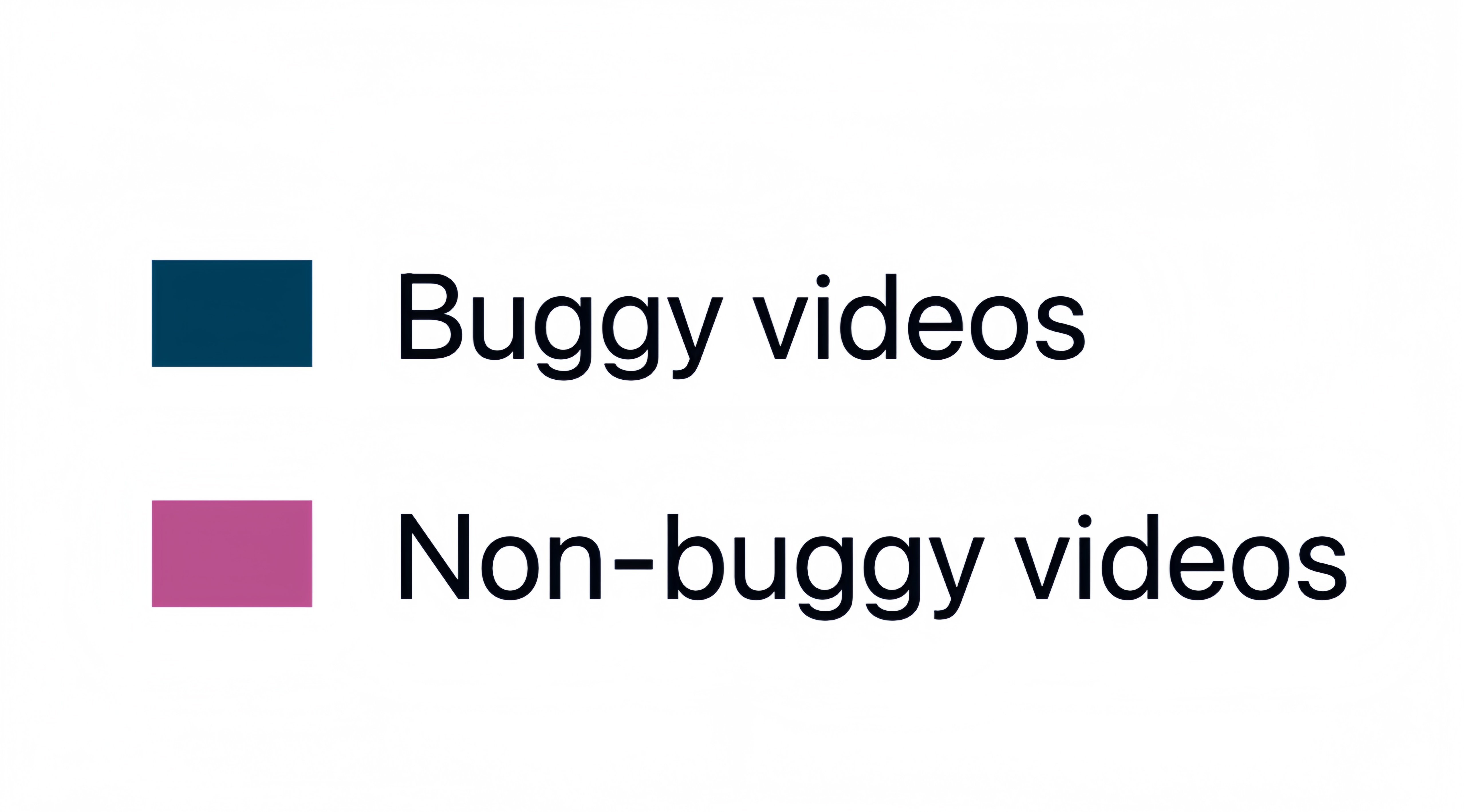}
    \end{minipage}
    
    \begin{minipage}[b]{\linewidth}
      \centering
      \includegraphics[width=\linewidth]{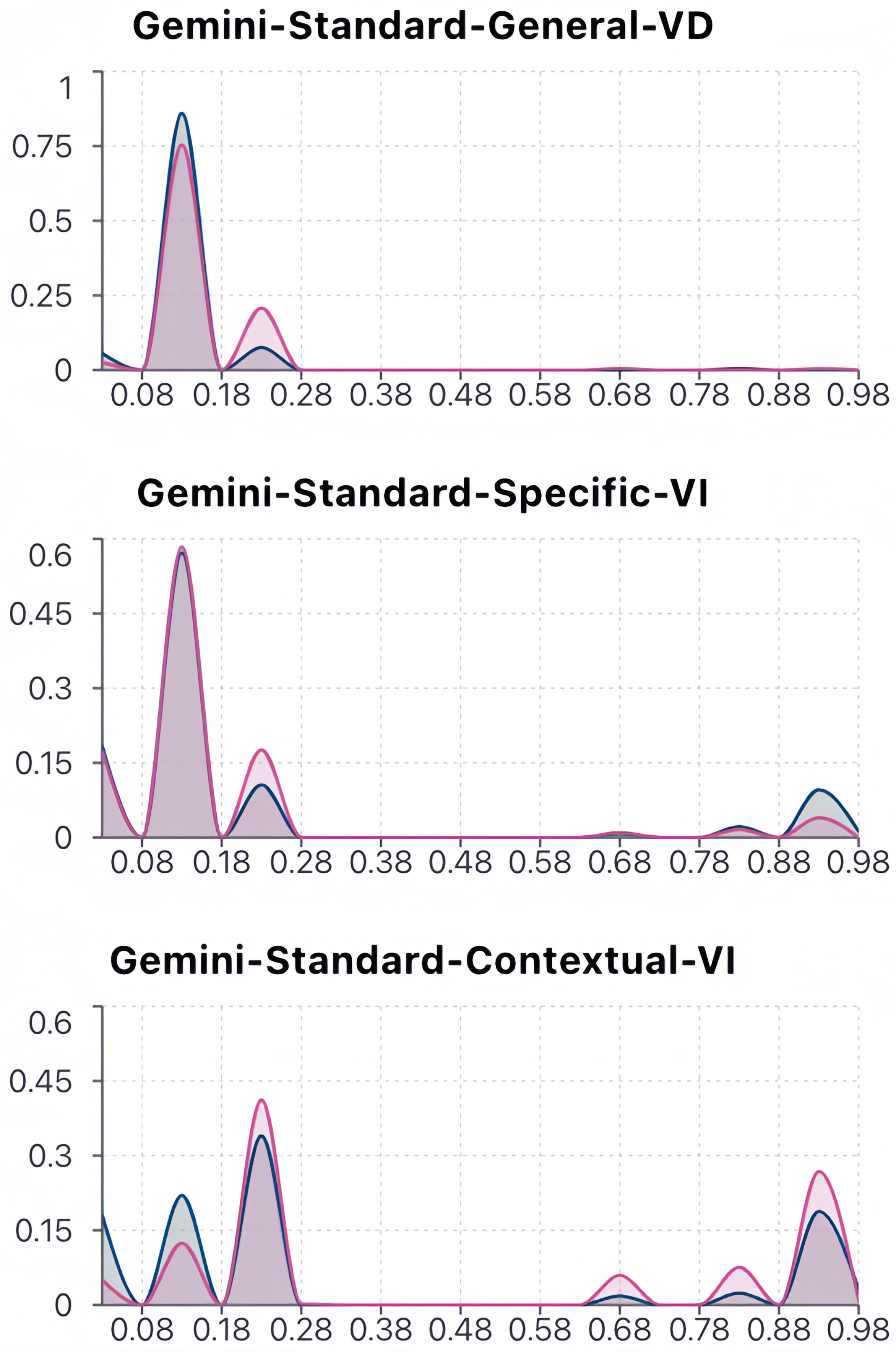}
      \caption{Gemini}
      \label{fig5}
    \end{minipage}
  \end{minipage}%
  
  \vspace{1em}

  \begin{minipage}[b]{0.245\linewidth}
    \centering
    \includegraphics[width=\linewidth]{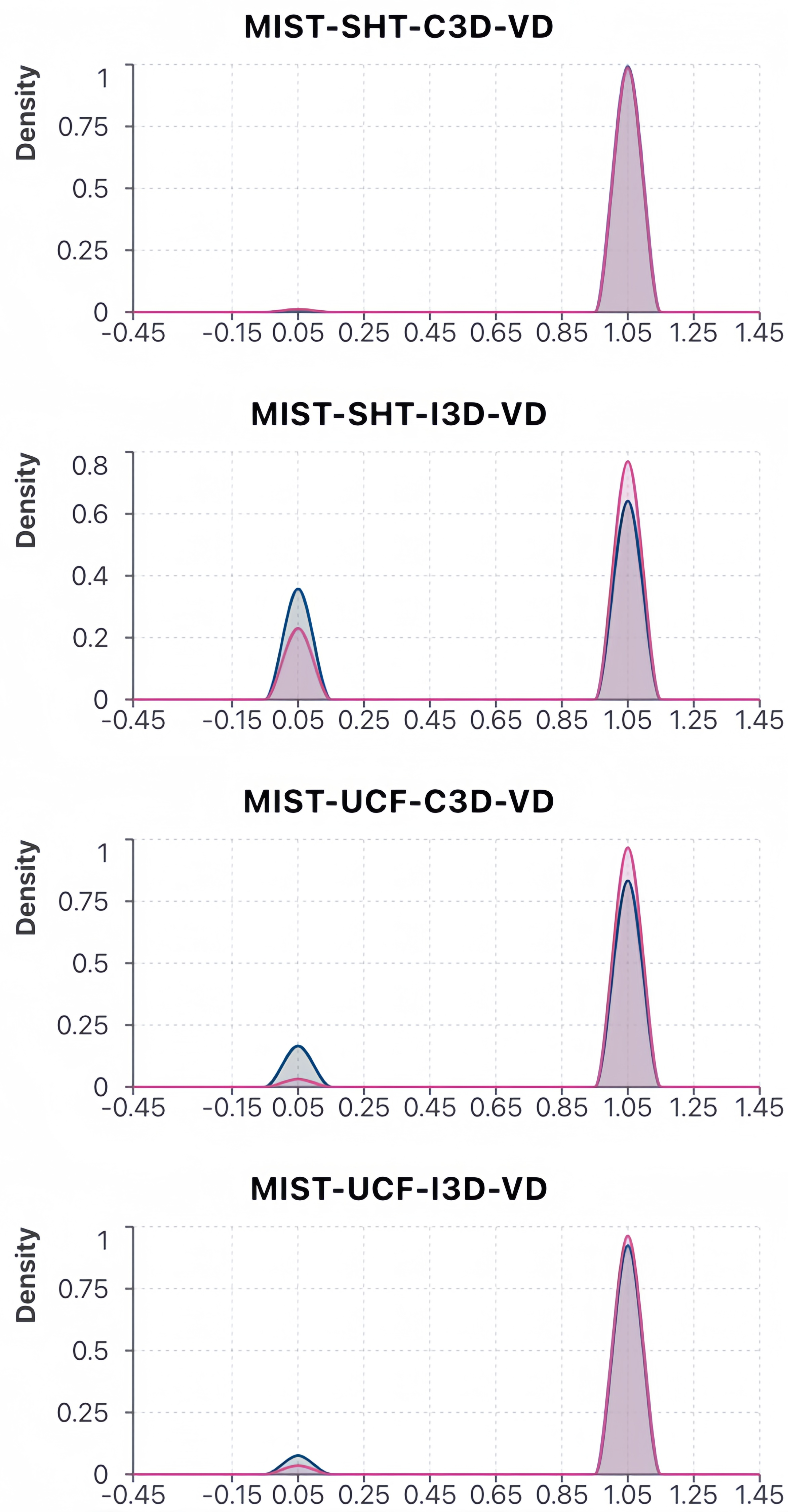}
    \caption{MIST}
    \label{fig6}
  \end{minipage}%
  \hfill%
  \begin{minipage}[b]{0.245\linewidth}
    \begin{minipage}[b]{\linewidth}
      \centering
      \includegraphics[width=\linewidth]{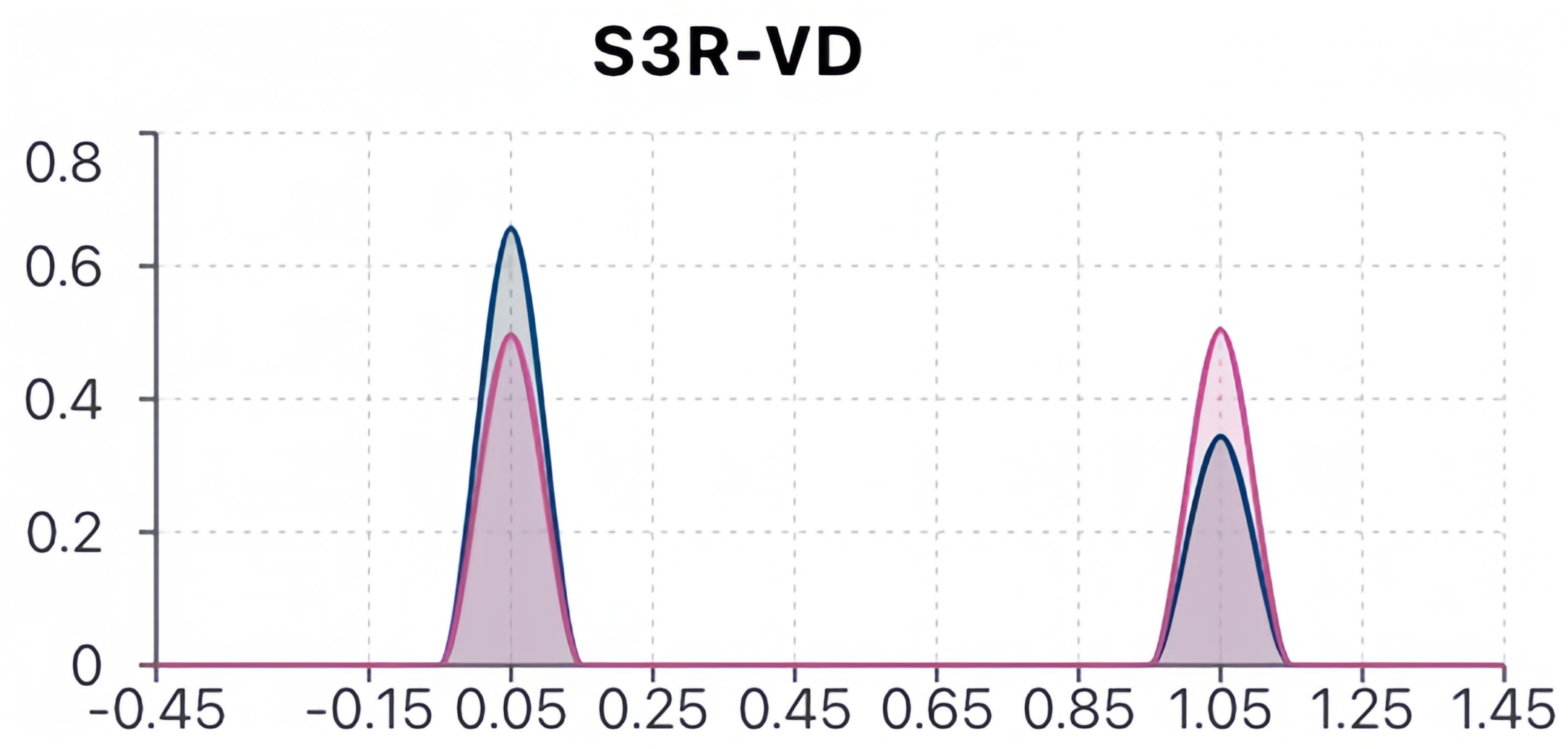}
      \caption{S3R}
      \label{fig7}
    \end{minipage}
    
    \begin{minipage}[b]{\linewidth}
      \centering
      \includegraphics[width=\linewidth]{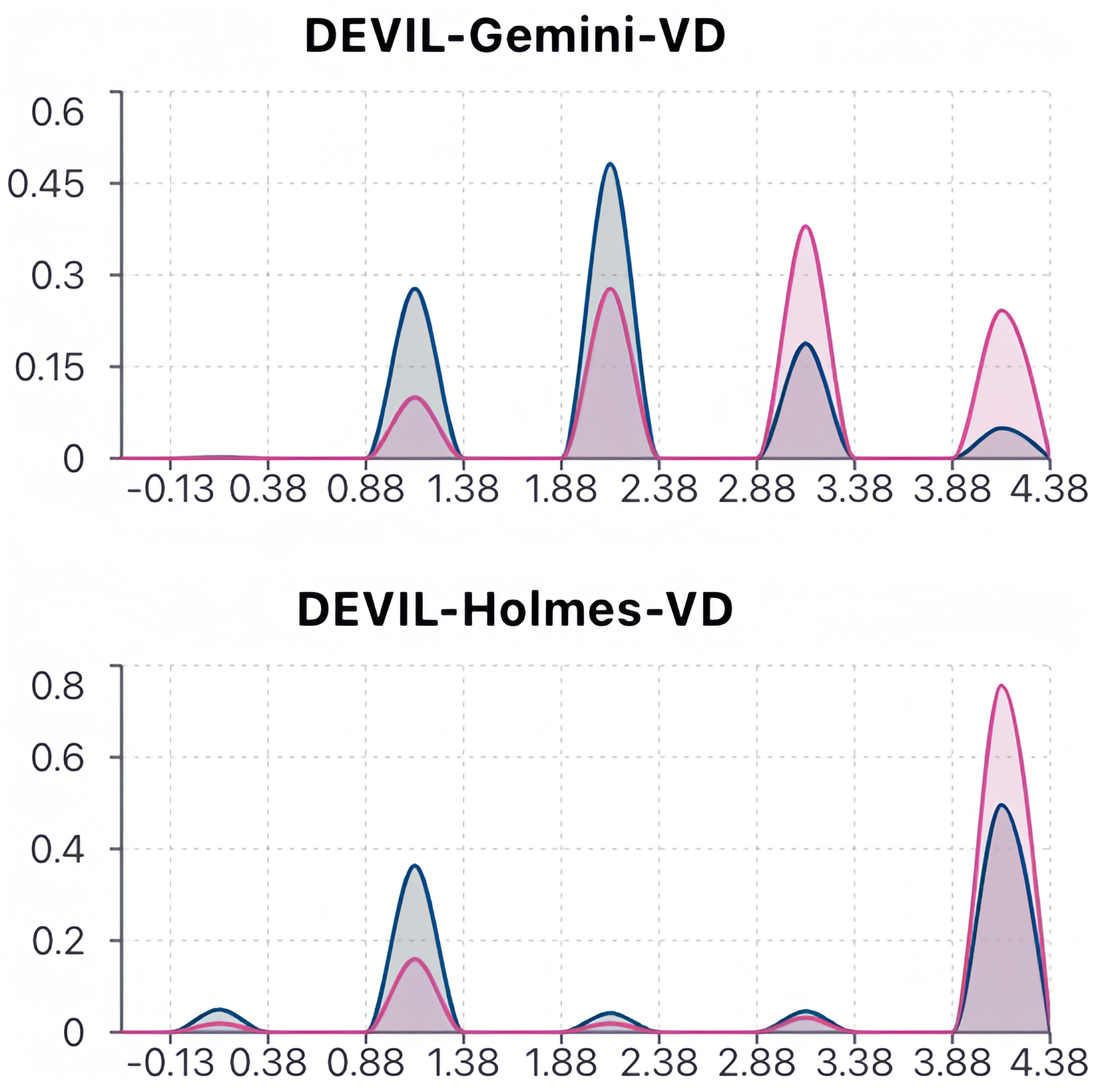}
      \caption{DEVIL}
      \label{fig8}
    \end{minipage}
  \end{minipage}%
  \hfill%
  \begin{minipage}[b]{0.245\linewidth}
    \centering
    \includegraphics[width=\linewidth]{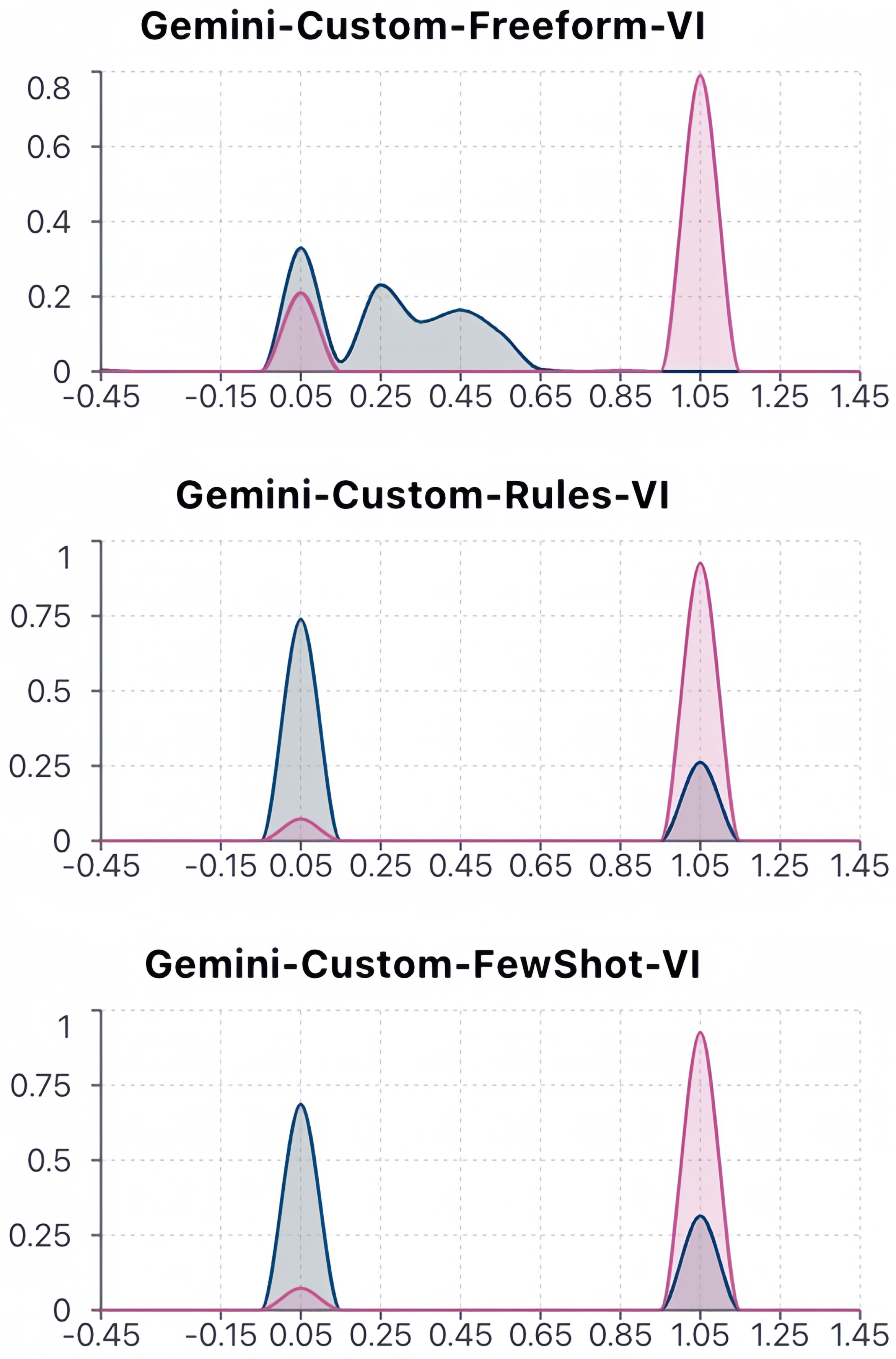}
    \caption{Gemini}
    \label{fig9}
  \end{minipage}%
  \hfill%
  \begin{minipage}[b]{0.245\linewidth}
    \centering
    \includegraphics[width=\linewidth]{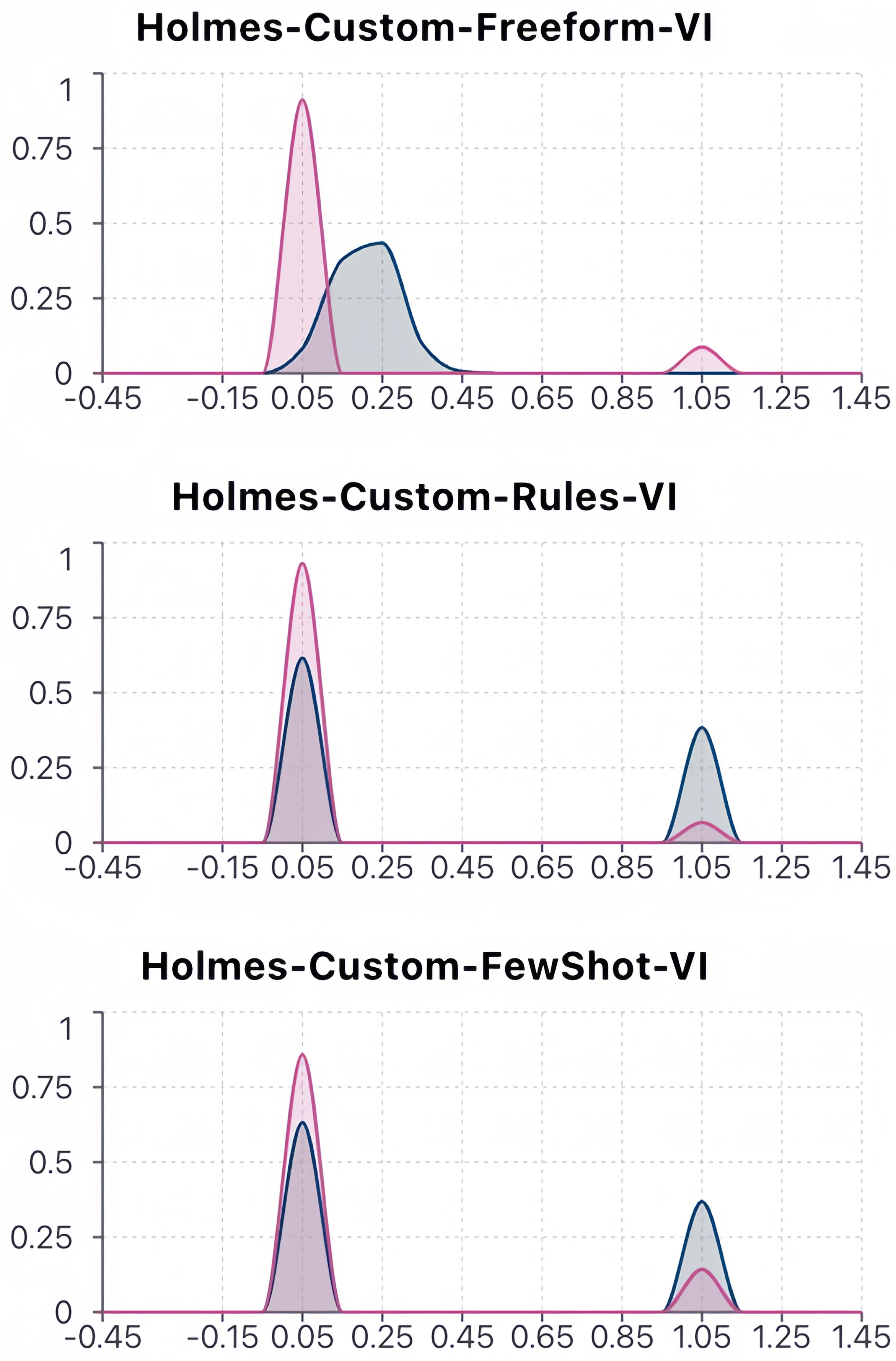}
    \caption{Holmes}
    \label{fig10}
  \end{minipage}%
  
  \caption{Score Distributions Across Approaches. \normalfont This figure presents the density plots of scores generated by different methods, illustrating their ability to distinguish between physics-compliant (pink) and physics-violating (blue) samples. The x-axis represents the model’s predicted score, where higher values indicate stronger adherence to physical laws. The y-axis represents the density of these scores. The ideal scenario is a clear separation between the two distributions, with the pink distribution (compliant samples) concentrated on the right and the blue distribution (violating samples) positioned on the left. Methods with significant overlap between distributions struggle to differentiate physics-compliant and violating samples, indicating higher classification uncertainty.}
  \label{fig11}
  \vspace{-1.5em}
\end{figure*}

\begin{figure*}[t!]
\centering
    \subfigure[F1 Score vs. AUC-ROC for VD (\physicsbench)]{
        \label{auc_f1_vd}
        \includegraphics[width=0.9\columnwidth]{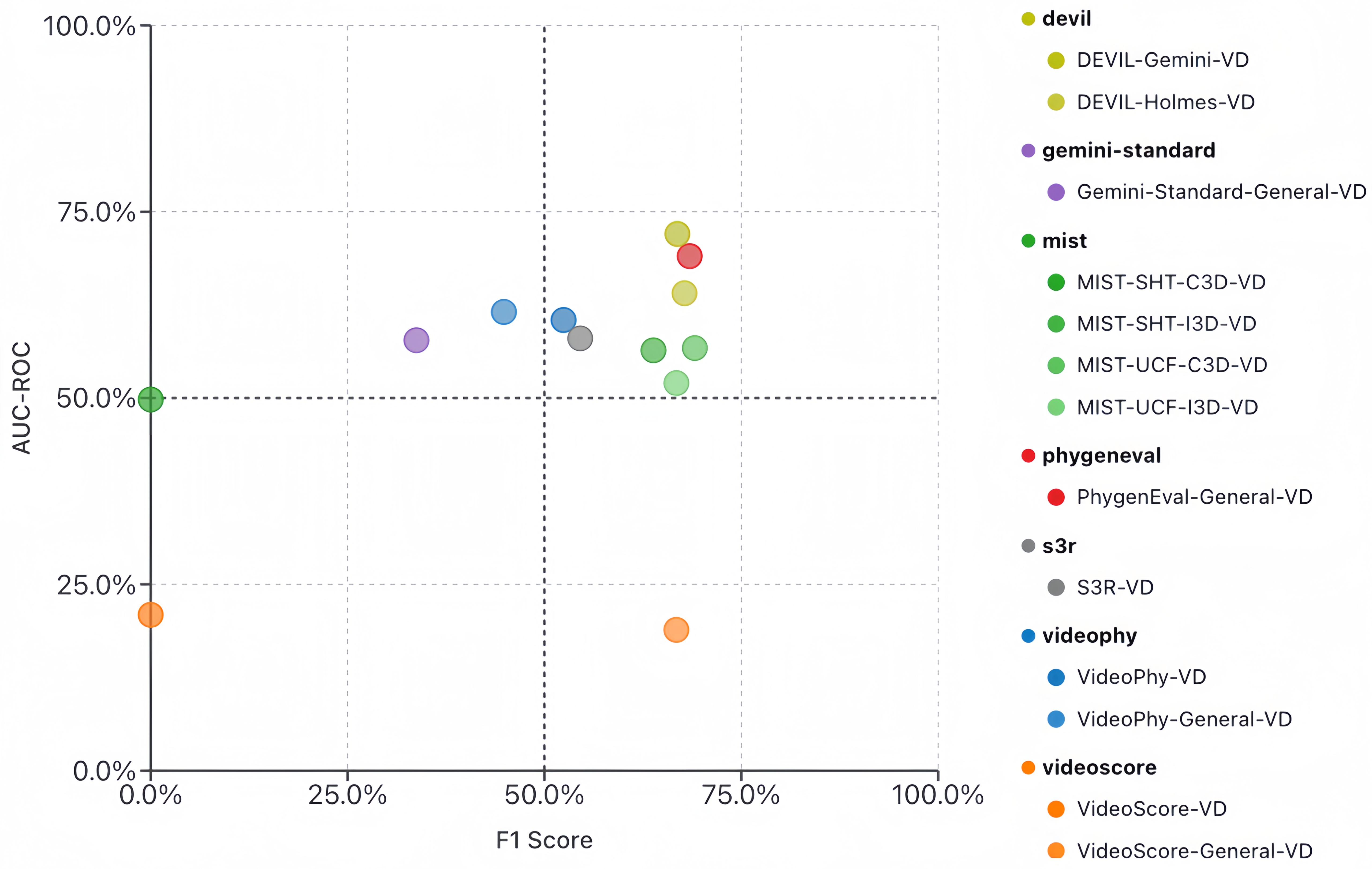}
    }
  \hfill%
    \subfigure[F1 Score vs. AUC-ROC for VI (\physicsbench)]{
        \label{auc_f1_vi}
        \includegraphics[width=0.9\columnwidth]{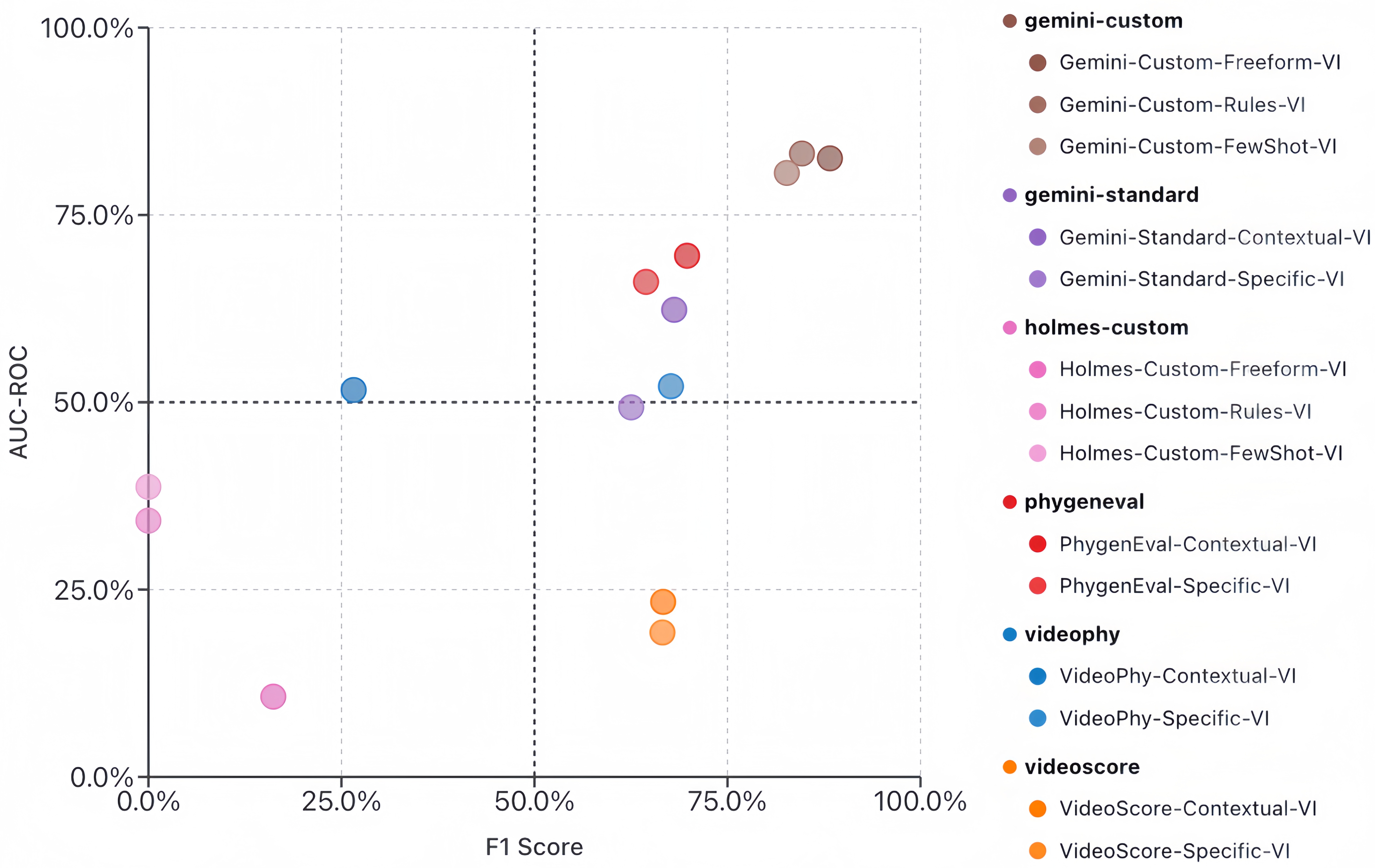}
    }  

    \subfigure[F1 Score vs. AUC-ROC for VD (\physicsbenchmulti)]{
        \label{auc_f1_vd_multi}
        \includegraphics[width=0.9\columnwidth]{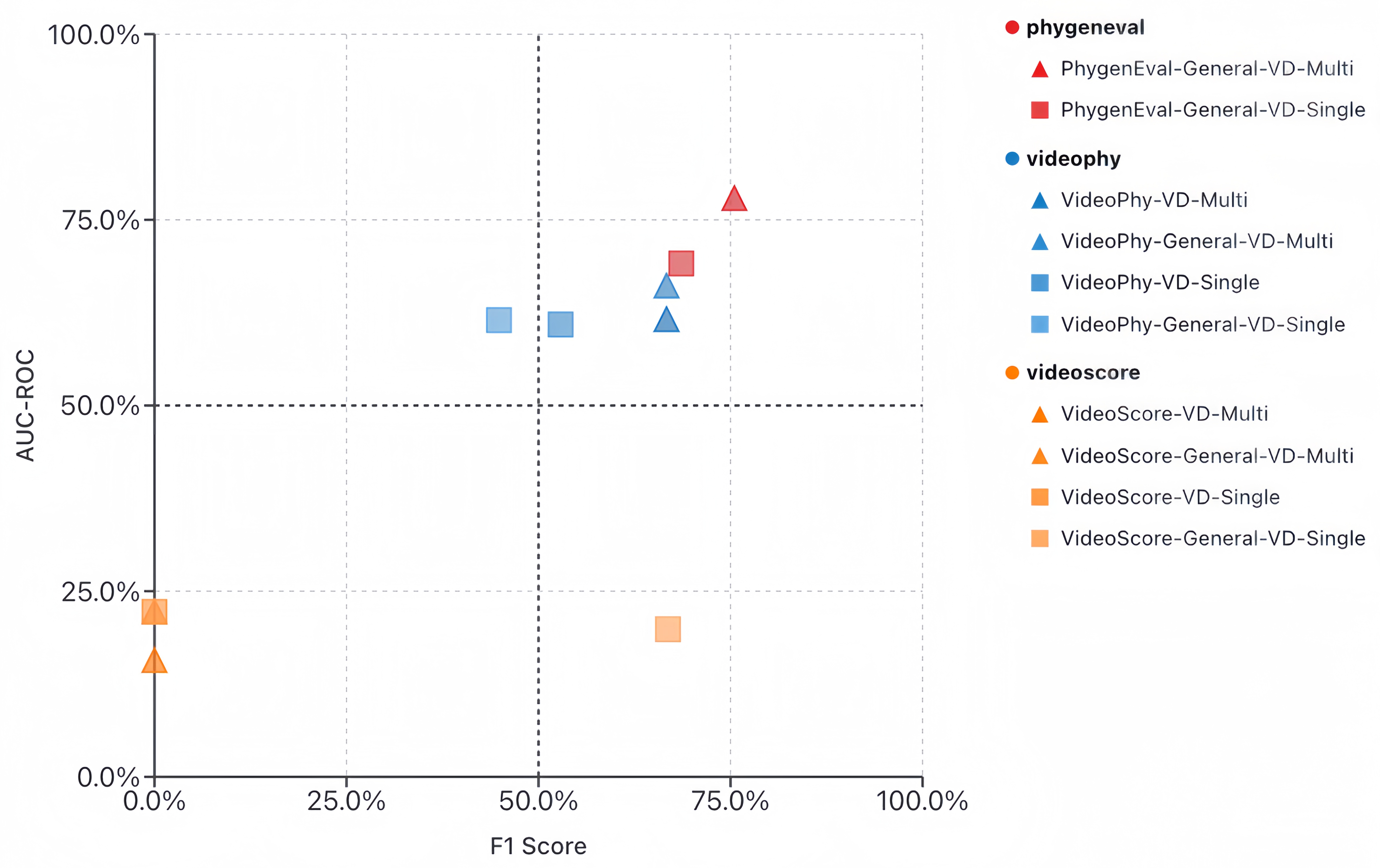}
    }
  \hfill%
    \subfigure[F1 Score vs. AUC-ROC for VI (\physicsbenchmulti)]{
        \label{auc_f1_vi_multi}
        \includegraphics[width=0.9\columnwidth]{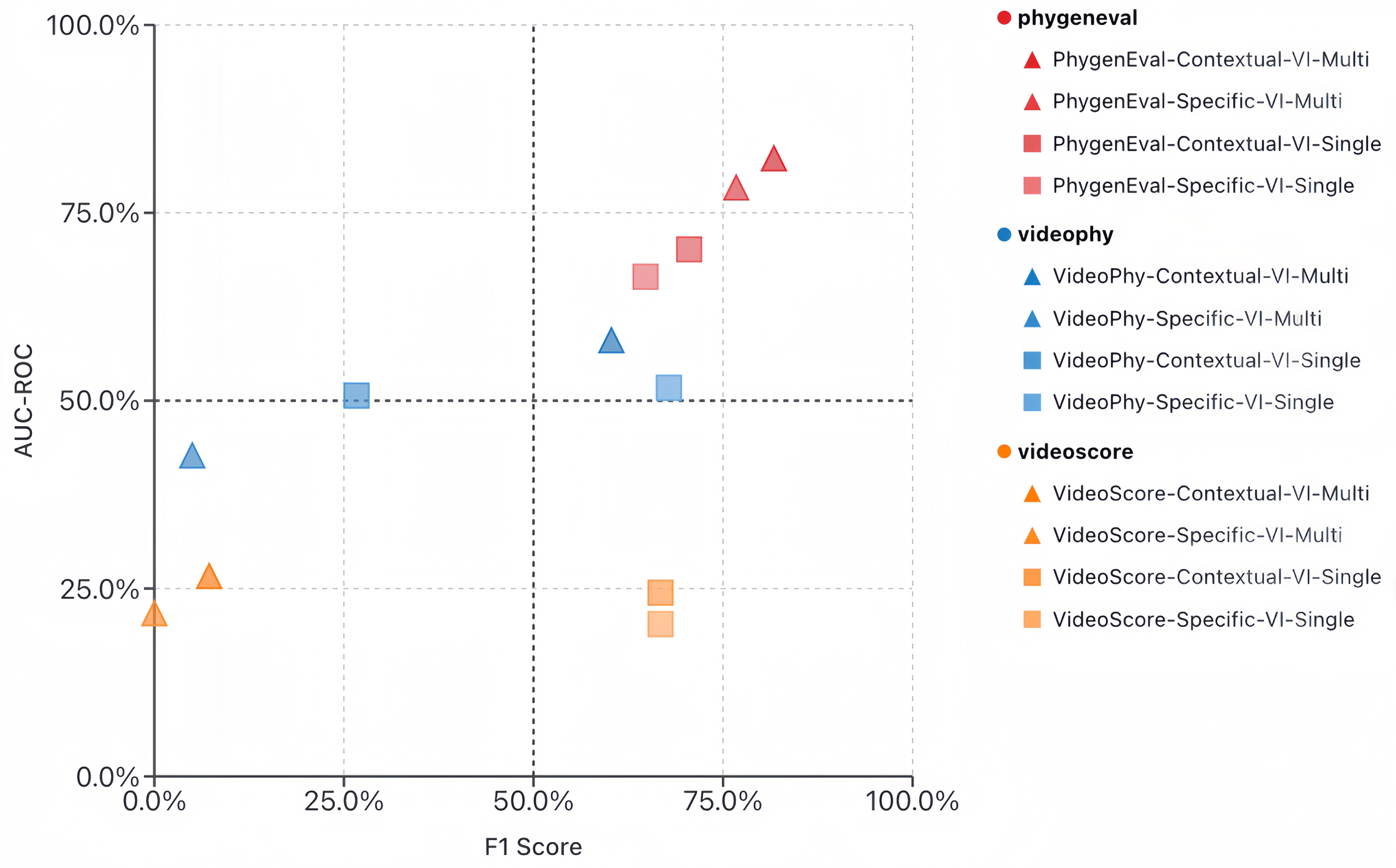}
    }  

  \caption{Performance Summary of Violation Detection (VD) and Violation Identification (VI) on \physicsbench. \normalfont This figure compares methods based on F1 Score (x-axis) and AUC-ROC (y-axis) for VD and VI tasks. Higher values indicate better performance, with the upper-right quadrant representing the most effective methods. The distribution highlights variations in detection and identification capabilities across approaches.}
  \label{auc_f1}
\end{figure*}

%% file: sections/results_analysis.tex
\begin{table}[t!]
\centering
\caption{Violation Detection (VD) Performance on \physicsbench}
\label{tab:vd_table}
\resizebox{\columnwidth}{!}{%
\begin{threeparttable}
\begin{tabular}{lcccccccc}
\toprule
\textbf{Model} & \textbf{A} & \textbf{P} & \textbf{R} & \textbf{F1} & \textbf{AUC} & \textbf{P.C.} & \textbf{S.C.} \\
\midrule
\rowcolor{mygray}
VideoPhy-VD & 57.9 & 60.3 & 46.4 & 52.4 & 60.5 & 19.0 & 18.1 \\
VideoPhy-General-VD & 57.7 & 64.4 & 34.4 & 44.9 & 61.5 & 22.0 & 20.0 \\
\midrule
\rowcolor{mygray}
VideoScore-VD & 50.0 & 0.0 & 0.0 & 0.0 & 20.9 & -47.0 & -50.5 \\
VideoScore-General-VD & 50.4 & 50.2 & \textbf{99.6} & 66.8 & 18.9 & -47.9 & -53.9 \\
\midrule
\rowcolor{mygray}
PhygenEval-General-VD & 67.0 & 65.6 & 71.6 & 68.5 & 69.0 & 36.0 & 36.4 \\
\midrule
MIST-SHT-C3D-VD & 50.0 & 0.0 & 0.0 & 0.0 & 49.8 & -2.0 & -2.0 \\
\rowcolor{mygray}
MIST-SHT-I3D-VD & 56.4 & 54.5 & 77.0 & 63.8 & 56.4 & 14.0 & 14.0 \\
MIST-UCF-C3D-VD & 56.7 & 53.7 & 96.8 & \textbf{69.1} & 56.7 & 22.4 & 22.4 \\
\rowcolor{mygray}
MIST-UCF-I3D-VD & 52.0 & 51.1 & 96.4 & 66.8 & 52.0 & 8.7 & 8.7 \\
\midrule
S3R-VD & 58.0 & 59.4 & 50.4 & 54.5 & 58.0 & 16.2 & 16.2 \\
\midrule
\rowcolor{mygray}
DEVIL-Gemini-VD & \textbf{69.2} & 72.3 & 62.2 & 66.9 & \textbf{72.0} & \textbf{39.7} & \textbf{39.9} \\
DEVIL-Holmes-VD & 63.5 & 60.7 & 76.8 & 67.8 & 64.1 & 28.0 & 28.4 \\
\midrule
\rowcolor{mygray}
Gemini-Standard-General-VD & 56.8 & \textbf{72.4} & 22.0 & 33.7 & 57.8 & 11.5 & 19.6 \\
\bottomrule
\end{tabular}%
\begin{tablenotes}
\footnotesize
\item All values in the table are multiplied by 100. \textbf{A}: Accuracy, \textbf{P}: Precision, \textbf{R}: Recall, \textbf{F1}: F1-Score, \textbf{AUC}: Area Under ROC Curve, \textbf{P.C.}: Pearson Correlation, \textbf{S.C.}: Spearman Correlation.
\end{tablenotes}
\end{threeparttable}
}
\end{table}

\begin{table}[t!]
\centering
\caption{Violation Identification (VI) Performance on \physicsbenchmulti}
\label{tab:vi_table}
\resizebox{\columnwidth}{!}{%
\begin{threeparttable}
\begin{tabular}{lcccccccc}
\toprule
\textbf{Model} & \textbf{A} & \textbf{P} & \textbf{R} & \textbf{F1} & \textbf{AUC} & \textbf{P.C.} & \textbf{S.C.} \\
\midrule
\rowcolor{mygray}
VideoPhy-Contextual-VI & 54.7 & 70.1 & 16.4 & 26.6 & 51.6 & 8.2 & 2.8 \\
VideoPhy-Specific-VI & 54.8 & 52.7 & 94.6 & 67.7 & 52.1 & 3.9 & 3.7 \\
\midrule
\rowcolor{mygray}
VideoScore-Contextual-VI & 50.3 & 50.2 & \textbf{99.4} & 66.7 & 23.4 & -44.5 & -46.2 \\
VideoScore-Specific-VI & 50.4 & 50.2 & 98.8 & 66.6 & 19.3 & -47.6 & -53.3 \\
\midrule
\rowcolor{mygray}
PhygenEval-Contextual-VI & 68.1 & 66.3 & 73.6 & 69.8 & 69.6 & 36.9 & 37.6 \\
PhygenEval-Specific-VI & 64.8 & 65.1 & 63.8 & 64.4 & 66.1 & 31.1 & 31.1 \\
\midrule
\rowcolor{mygray}
Gemini-Standard-Contextual-VI & 61.3 & 57.9 & 82.6 & 68.1 & 62.3 & 18.0 & 22.2 \\
Gemini-Standard-Specific-VI & 50.5 & 50.3 & 82.6 & 62.5 & 49.3 & -10.5 & -1.3 \\
\midrule
\rowcolor{mygray}
Gemini-Custom-Freeform-VI & \textbf{89.5} & \textbf{100.0} & 79.0 & \textbf{88.3} & 82.5 & 64.3 & 58.8 \\
Gemini-Custom-Rules-VI & 83.2 & 77.9 & 92.6 & 84.6 & \textbf{83.2} & \textbf{67.6} & \textbf{67.6} \\
\rowcolor{mygray}
Gemini-Custom-FewShot-VI & 80.6 & 74.7 & 92.6 & 82.7 & 80.6 & 63.0 & 63.0 \\
\midrule
Holmes-Custom-Freeform-VI & 54.4 & \textbf{100.0} & 8.8 & 16.2 & 10.7 & -26.9 & -72.1 \\
\rowcolor{mygray}
Holmes-Custom-Rules-VI & 50.0 & 0.0 & 0.0 & 0.0 & 34.2 & -37.8 & -37.8 \\
Holmes-Custom-FewShot-VI & 50.0 & 0.0 & 0.0 & 0.0 & 38.7 & -25.9 & -25.9 \\
\bottomrule
\end{tabular}%
\begin{tablenotes}[flushleft]\footnotesize
\item All values in the table are multiplied by 100. \textbf{A}: Accuracy, \textbf{P}: Precision, \textbf{R}: Recall, \textbf{F1}: F1-Score, \textbf{AUC}: Area Under ROC Curve, \textbf{P.C.}: Pearson Correlation, \textbf{S.C.}: Spearman Correlation.
\end{tablenotes}
\end{threeparttable}}
\vspace{-1em}
\end{table}

\begin{table}[t!]
\centering
\caption{Violation Detection (VD) Performance on \physicsbenchmulti}
\label{tab:vd_table}
\resizebox{\columnwidth}{!}{%
\begin{threeparttable}
\begin{tabular}{lcccccccc}
\toprule
\textbf{Model} & \textbf{A} & \textbf{P} & \textbf{R} & \textbf{F1} & \textbf{AUC} & \textbf{P.C.} & \textbf{S.C.} \\
\midrule
\rowcolor{mygray} \textbf{VideoPhy-VD-Multi} & 61.5 & 58.8 & 76.9 & 66.7 & 61.6 & 19.0 & 20.1 \\
VideoPhy-VD-Single & 58.2 & 60.7 & 46.9 & 52.9 & 60.9 & 19.7 & 18.9 \\
\midrule
\rowcolor{mygray} \textbf{VideoPhy-General-VD-Multi} & 62.8 & 60.4 & 74.4 & 66.7 & 66.1 & 27.9 & 28.0 \\
VideoPhy-General-VD-Single & 57.6 & 64.1 & 34.5 & 44.9 & 61.5 & 21.8 & 19.9 \\
\midrule
\rowcolor{mygray} \textbf{VideoScore-VD-Multi} & 50.0 & 0.0 & 0.0 & 0.0 & 15.7 & -48.1 & -59.6 \\
VideoScore-VD-Single & 50.0 & 0.0 & 0.0 & 0.0 & 22.2 & -46.0 & -48.2 \\
\midrule
\rowcolor{mygray} \textbf{VideoScore-General-VD-Multi} & 50.0 & 0.0 & 0.0 & 0.0 & 22.3 & -48.4 & -48.1 \\
VideoScore-General-VD-Single & 50.5 & 50.3 & \textbf{99.8} & 66.9 & 19.9 & -46.5 & -52.2 \\
\midrule
\rowcolor{mygray} \textbf{PhygenEval-General-VD-Multi} & \textbf{76.3} & \textbf{78.1} & 73.1 & \textbf{75.5} & \textbf{77.9} & \textbf{52.8} & \textbf{54.0} \\
PhygenEval-General-VD-Single & 66.9 & 65.3 & 72.2 & 68.6 & 69.1 & 36.2 & 36.5 \\
\bottomrule
\end{tabular}%
\begin{tablenotes}
\footnotesize
\item All values in the table are multiplied by 100. \textbf{A}: Accuracy, \textbf{P}: Precision, \textbf{R}: Recall, \textbf{F1}: F1-Score, \textbf{AUC}: Area Under ROC Curve, \textbf{P.C.}: Pearson Correlation, \textbf{S.C.}: Spearman Correlation.
\end{tablenotes}
\end{threeparttable}
}
\end{table}

\begin{table}[t!]
\centering
\caption{Visual Inference (VI) Performance on \physicsbenchmulti}
\label{tab:vi_table}
\resizebox{\columnwidth}{!}{%
\begin{threeparttable}
\begin{tabular}{lcccccccc}
\toprule
\textbf{Model} & \textbf{A} & \textbf{P} & \textbf{R} & \textbf{F1} & \textbf{AUC} & \textbf{P.C.} & \textbf{S.C.} \\
\midrule
\rowcolor{mygray} \textbf{VideoPhy-Contextual-VI-Multi} & 62.8 & 64.7 & 56.4 & 60.3 & 58.0 & 17.0 & 13.9 \\
VideoPhy-Contextual-VI-Single & 54.7 & 69.7 & 16.5 & 26.7 & 50.7 & 6.4 & 1.2 \\
\midrule
\rowcolor{mygray} \textbf{VideoPhy-Specific-VI-Multi} & 51.3 & \textbf{100.0} & 2.6 & 5.0 & 42.7 & -13.0 & -12.6 \\
VideoPhy-Specific-VI-Single & 55.2 & 52.9 & 94.6 & 67.9 & 51.7 & 3.5 & 3.0 \\
\midrule
\rowcolor{mygray} \textbf{VideoScore-Contextual-VI-Multi} & 50.6 & 60.0 & 3.8 & 7.2 & 26.7 & -44.4 & -40.4 \\
VideoScore-Contextual-VI-Single & 50.4 & 50.2 & 99.6 & 66.8 & 24.5 & -43.1 & -44.3 \\
\midrule
\rowcolor{mygray} \textbf{VideoScore-Specific-VI-Multi} & 50.0 & 0.0 & 0.0 & 0.0 & 21.7 & -48.9 & -49.1 \\
VideoScore-Specific-VI-Single & 50.7 & 50.3 & \textbf{99.1} & 66.8 & 20.3 & -46.2 & -51.5 \\
\midrule
\rowcolor{mygray} \textbf{PhygenEval-Contextual-VI-Multi} & \textbf{80.8} & 77.9 & 85.9 & \textbf{81.7} & \textbf{82.2} & \textbf{60.1} & \textbf{62.0} \\
PhygenEval-Contextual-VI-Single & 68.5 & 66.3 & 75.3 & 70.5 & 70.1 & 38.1 & 38.8 \\
\midrule
\rowcolor{mygray} \textbf{PhygenEval-Specific-VI-Multi} & 76.3 & 75.3 & 78.2 & 76.7 & 78.4 & 53.1 & 54.2 \\
PhygenEval-Specific-VI-Single & 65.0 & 65.1 & 64.4 & 64.8 & 66.5 & 32.0 & 31.9 \\
\bottomrule
\end{tabular}%
\begin{tablenotes}
\footnotesize
\item All values in the table are multiplied by 100. \textbf{A}: Accuracy, \textbf{P}: Precision, \textbf{R}: Recall, \textbf{F1}: F1-Score, \textbf{AUC}: Area Under ROC Curve, \textbf{P.C.}: Pearson Correlation, \textbf{S.C.}: Spearman Correlation.
\end{tablenotes}
\end{threeparttable}
}
\end{table}

\section{RQ2: Results and Analysis}
\label{sec:rq2-results}

In this section, we present the results of our experimental evaluation addressing RQ2. We systematically analyze the performance of various approaches across both violation detection and violation identification capabilities.

\subsection{Overall Performance Comparison}
\label{sec:overall-performance}

Tables~\ref{tab:vd_table} and~\ref{tab:vi_table} present the quantitative evaluation of different approaches for \textbf{Violation Detection} (VD) and \textbf{Violation Identification} (VI), respectively. Violation Detection measures whether an approach can correctly determine if a video violates any physical rules, while Violation Identification assesses whether an approach can accurately identify which specific physical rule is being violated. Figure~\ref{auc_f1} visualizes the relationship between F1-score and AUC-ROC metric for each approach, providing an intuitive comparison of their effectiveness.

\subsubsection{Violation Detection Performance}

For violation detection (Table~\ref{tab:vd_table}), we observe substantial variation in performance across approaches. The highest accuracy is achieved by \texttt{DEVIL-Gemini-VD} at 69.2\%, with an impressive precision of 72.3\% and the highest AUC of 72.0\%. This approach also demonstrates the strongest correlation with human judgments, as indicated by Pearson (39.7\%) and Spearman (39.9\%) correlation coefficients. \texttt{PhygenEval-General-VD} follows with 67.0\% accuracy and balanced precision (65.6\%) and recall (71.6\%), resulting in a solid F1-score of 68.5\%.

In contrast, certain approaches show significant limitations. \texttt{\allowbreak VideoScore-VD} fails to detect any violations, with 0\% precision and recall, exhibiting strong negative correlations with human judgments (-47.0\% Pearson, -50.5\% Spearman). Similarly, \texttt{MIST-SHT-C3D-\allowbreak VD} performs at chance level accuracy (50.0\%) with no true positive detections.

Looking at traditional video anomaly detection approaches, we observe that MIST-VAD variants trained on UCF-Crime dataset demonstrate reasonable performance, with \texttt{MIST-UCF-C3D-VD} achieving the highest F1-score overall at 69.1\%, albeit with lower precision (53.7\%) due to its extremely high recall (96.8\%). This suggests that these models tend to over-classify videos as containing physics violations.

Our analysis reveals that LMM-based methods (Gemini, Holmes, PhyGenEval) generally outperform traditional computer vision methods. Further statistical analysis shows a correlation coefficient of 0.42 between F1 and AUC-ROC metrics, and 0.29 with Pearson correlation, indicating that models with higher F1 and AUC-ROC scores tend to better align with human judgments. Comparing \texttt{DEVIL-Gemini-VD} and \texttt{Gemini-Standard-General-VD}, we find that prompt engineering significantly impacts performance, with DEVIL's five-level naturalness rating approach proving more effective than simple binary classification prompts.

\subsubsection{Violation Identification Performance}

For violation identification (Table~\ref{tab:vi_table}), the results reveal a clear advantage of specialized prompt-based approaches using large multimodal models. The custom Gemini prompts (\texttt{Gemini-\allowbreak Custom-Freeform-VI}, \texttt{Gemini-\allowbreak Custom-Rules-VI}, \texttt{Gemini-Custom-FewShot-VI}) significantly outperform other approaches, with \texttt{Gemini-Custom-Freeform-VI} achieving 89.5\% accuracy and perfect precision (100\%), yielding an F1-score of 88.3\%. These approaches also demonstrate much stronger correlations with human judgments, with \texttt{Gemini-Custom-Rules-VI} achieving 67.6\% for both Pearson and Spearman correlations.

In stark contrast, the Holmes-based variants (\texttt{Holmes-Custom-\allowbreak Freeform-VI}, \texttt{Holmes-Custom-Rules-VI}, \texttt{Holmes-Custom-\allowbreak FewShot-\allowbreak VI}) struggle with violation identification, with \texttt{Holmes-Custom-Rules-\allowbreak VI} and \texttt{Holmes-Custom-FewShot-VI} failing to identify any specific violations (0\% precision and recall). The performance gap between Gemini and Holmes variants highlights the critical importance of underlying model capabilities in understanding specific physical principles being violated.

\texttt{VideoScore-VI} approaches maintain extremely high recall (98.8-99.4\%) but at the cost of precision, resulting in F1-scores around 66-67\%. This suggests these models are biased toward classifying most inputs as violations of any given rule, limiting their practical utility for specific violation identification. Their negative correlation values further indicate that their judgments often contradict human assessments.

Across the contextual and specific prompt approaches, we observe that prompts including video content descriptions (\texttt{*-Contextual-\allowbreak VI}) generally outperform prompts that only describe specific physical rules (\texttt{*-Specific-VI}). This indicates that providing context about the video content helps models better understand and apply physical rules. Interestingly, Gemini's performance with standard rules-based prompts exceeds its performance with few-shot examples, suggesting that few-shot examples may not always improve physics rule understanding for these models.

Comparing Holmes and Gemini, Holmes shows moderate violation detection but fails in precise violation identification, while Gemini excels at both detection and specific identification, indicating a better understanding of physical principles.

\subsection{Detailed Analysis of Score Distributions}
\label{sec:distribution-analysis}

Figure~\ref{fig11} shows the score distributions for various approaches, highlighting patterns in how these methods differentiate between physics-compliant and physics-violating samples.

\subsubsection{Distribution Analysis for Physics-Centric Approaches}

The density plots in Figure~\ref{fig11} expose important characteristics of each approach's discrimination ability:

    \textbf{VideoPhy} (Figure~\ref{fig2}) shows moderate separation between distributions for compliant (pink) and violating (blue) samples, with multiple overlapping peaks suggesting areas of uncertainty in its decision boundary.

    \textbf{VideoScore} (Figure~\ref{fig3}) exhibits poor separation, with distributions heavily overlapping and notably, the pink distribution (compliant videos) leaning toward the left side. This leftward bias indicates judgments contrary to ground truth, explaining its negative Pearson and Spearman correlations and overall weak discriminative performance.

    \textbf{PhyGenEval} (Figure~\ref{fig4}) demonstrates clearer bimodal distributions with the compliant (pink) distribution leaning more toward the right side, showing better separation. However, most results concentrate near central peaks rather than at extremes, suggesting moderate confidence in classifications despite its stronger quantitative performance in Table~\ref{tab:vd_table}.

    \textbf{Gemini Baseline} (Figure~\ref{fig5}) shows multiple peaks with partial overlap, suggesting distinct classification patterns for different types of physics violations, but with regions of uncertainty.

\subsubsection{Distribution Analysis for Video Anomaly and LMM Approaches}

The distributions in Figure~\ref{fig11} offer additional insights:

    \textbf{MIST-VAD variants} (Figure~\ref{fig6}) show significant overlap between compliant and violating distributions, especially in subfigures 1 and 4, indicating weaker classification ability and explaining their inconsistent performance.

    \textbf{S3R} (Figure~\ref{fig7}) displays moderate separation with overlapping regions, consistent with its balanced precision-recall performance in violation detection.

    \textbf{DEVIL-integrated approaches} (Figures~\ref{fig8}-\ref{fig10}) display distinct multimodal distributions. Compliant distributions (pink) are skewed to the right with minimal overlap, indicating strong classification. \texttt{DEVIL-Gemini} shows multiple peaks, suggesting specialized detection of various physics violation levels.

    \textbf{Gemini-cus and Holmes-cus} demonstrate contrasting behaviors despite both being LLMs. Gemini-cus shows minimal overlap between compliant and violating distributions, with compliant samples (pink) clearly positioned toward the right side, establishing a sharp decision boundary and exhibiting the strongest overall performance. Conversely, Holmes-cus shows compliant distributions leaning toward the left side, suggesting classifications contrary to ground truth.

This distribution analysis helps explain why certain approaches achieve high recall but suffer in precision (e.g., MIST-VAD and VideoScore), while others maintain better balanced performance (PhyGenEval, Gemini-based approaches). The rightward shift of compliant distributions generally indicates better alignment with ground truth, while leftward shifts suggest inverse correlations with actual physics compliance.

\subsection{Comparative Analysis Across Approaches}
\label{sec:comparative-analysis}

To understand broader patterns, we analyze performance across the four major categories of approaches:

\subsubsection{Deep Learning Based Video Evaluation}
Traditional video anomaly detection approaches (MIST-VAD and S3R) demonstrate reasonable performance for violation detection but with important limitations. These approaches excel at detecting obvious anomalies in motion patterns, with UCF-trained variants achieving some of the highest recall values (96.4-96.8\%). However, their precision remains moderate (51.1-59.4\%), indicating a tendency to over-classify normal physics behaviors as anomalous.
The feature encoder choice significantly impacts performance, with I3D generally outperforming C3D for the ShanghaiTech dataset but showing comparable or slightly worse performance for the UCF-Crime dataset. This suggests that the dataset characteristics interact with feature extraction methods in complex ways for physics violation detection.
These approaches are limited by their inability to identify specific violations, as they operate solely at the motion pattern level without considering textual descriptions or physical rule understanding.

\subsubsection{Pure Prompt Engineering with General-purpose LMMs}

Approaches leveraging prompt engineering with general-purpose LMMs (\texttt{Gemini-Standard} and DEVIL-integrated approaches) show promising results. \texttt{DEVIL-\allowbreak Gemini-VD} achieves the highest accuracy (69.2\%) and AUC (72.0\%) for violation detection, demonstrating that carefully designed prompts can effectively leverage general-purpose multimodal capabilities.
The performance difference between \texttt{Gemini-Standard-General-\allowbreak VD} (56.8\% accuracy) and \texttt{DEVIL-\allowbreak Gemini-\allowbreak VD} (69.2\% accuracy) highlights the critical importance of prompt design. The DEVIL-integrated approach, with its structured evaluation protocol focusing on temporal consistency and dynamics assessment, provides more effective guidance for the model's physics violation assessment.

\subsubsection{Fine-tuned Multimodal Models}

VideoScore and VideoPhy, representing fine-tuned multimodal approaches, demonstrate mixed results. VideoPhy shows moderate performance for violation detection (57.9\% accuracy) but struggles with consistent violation identification. VideoScore performs poorly for violation detection (50.0\% accuracy with 0\% precision) despite maintaining high recall for violation identification.
The distribution analysis reveals that these models often lack clear separation between compliant and violating samples, suggesting limitations in their fine-tuning objectives or training data for physics-specific evaluation tasks.

\subsubsection{Physics-Centric Fine-tuning}

PhyGenEval (\texttt{PhygenEval-VD}) demonstrates strong and balanced performance across metrics (67.0\% accuracy, 65.6\% precision, 71.6\% recall) for violation detection. Its hierarchical evaluation framework, specifically designed for physics assessment, provides effective discrimination between compliant and violating behaviors. 
When adapted for violation identification (\texttt{PhygenEval-Contextual-\allowbreak VI} and \texttt{PhygenEval-Specific-VI}), it maintains solid performance (68.1\% and 64.8\% accuracy), though not matching the custom Gemini prompts. This suggests that physics-centric fine-tuning provides robust foundations for violation detection and identification tasks.

\subsection{Performance on Multi-Violation Scenarios}

To further investigate the robustness of detection approaches, we analyze their performance on \physicsbenchmulti, a challenging subset containing 39 videos where multiple \physicsbugs occur simultaneously. This subset represents particularly complex scenarios that may pose additional challenges for detection algorithms due to the interaction and potential masking effects of concurrent violations.

\subsubsection{Comparative Analysis: Multi-Violation vs. Single-Violation Detection}

Tables~\ref{tab:vd_table} and~\ref{tab:vi_table} present the performance metrics for violation detection (VD) and violation identification (VI) on \physicsbenchmulti. Surprisingly, our results reveal that several approaches demonstrate \emph{improved} performance on multi-violation scenarios compared to the full \physicsbench dataset.

For violation detection, \textbf{PhygenEval-General-VD} shows the most notable improvement, achieving 76.3\% accuracy on \physicsbenchmulti compared to 67.0\% on \physicsbench—a remarkable 9.3 percentage point increase. The AUC metric similarly improves from 69.0\% to 77.9\%, while both correlation metrics show substantial gains (Pearson: 36.0\% → 52.8\%, Spearman: 36.4\% → 54.0\%). This counter-intuitive finding suggests that the presence of multiple concurrent violations may actually create more pronounced deviations from expected physical behavior, making detection easier for physics-aware models.

\textbf{VideoPhy} variants exhibit mixed results. While VideoPhy-VD shows a modest improvement in accuracy (57.9\% → 61.5\%), VideoPhy-General-VD demonstrates more substantial gains across multiple metrics, particularly in recall (34.4\% → 74.4\%) and correlation coefficients. This pattern indicates that general physics understanding becomes more valuable when multiple violations are present, as the cumulative effect of violations may create clearer signals for detection.

In stark contrast, \textbf{VideoScore} approaches continue to struggle with multi-violation scenarios. VideoScore-VD maintains 0\% precision and recall on both datasets, while its correlation metrics worsen on \physicsbenchmulti (Pearson: -47.0\% → -48.1\%, Spearman: -50.5\% → -59.6\%). This deterioration suggests that VideoScore's underlying model lacks the capability to handle complex physics interactions, with performance degrading further as scenario complexity increases.

\subsubsection{Violation Identification in Multi-Bug Contexts}

For violation identification tasks (Table~\ref{tab:vi_table}), the performance patterns become more nuanced. \textbf{PhygenEval-Contextual-VI} achieves exceptional performance on \physicsbenchmulti with 80.8\% accuracy and 81.7\% F1-score, representing improvements of 12.7 and 11.9 percentage points respectively compared to \physicsbench. The correlation metrics show even more dramatic improvements, with Pearson correlation increasing from 36.9\% to 60.1\% and Spearman from 37.6\% to 62.0\%.

Interestingly, the contextual variants consistently outperform their specific counterparts on multi-violation scenarios. For instance, VideoPhy-Contextual-VI achieves 62.8\% accuracy on \physicsbenchmulti compared to 54.7\% on \physicsbench, while VideoPhy-Specific-VI shows a slight decrease (54.8\% → 51.3\%). This pattern suggests that contextual information becomes increasingly important when multiple physics principles are violated simultaneously, as the model needs to understand the broader scene context to disentangle concurrent violations.

\subsection{Key Insights and Implications}
\label{sec:key-insights}

Our comprehensive evaluation yields several important insights for \physicsbug detection:
    \textbf{Prompt Engineering Effectiveness:} Custom-designed prompts with general-purpose LMMs (particularly Gemini) outperform many specialized fine-tuned approaches, especially for violation identification tasks. This suggests that existing LMMs already possess substantial physics understanding that can be effectively elicited through careful prompting.

\textbf{Precision-Recall Trade-offs:} Many approaches exhibit extreme precision-recall trade-offs, either detecting most violations but with many false positives (high recall, low precision) or being highly selective but missing many violations (high precision, low recall). The most practically useful approaches (PhyGenEval, DEVIL-Gemini) maintain better balance.

\textbf{Model Capability Gaps:} The stark performance difference between Gemini and Holmes approaches for violation identification tasks (88.3\% vs. 0\% F1-score) highlights significant capability gaps between different LMMs for physics understanding, despite both being recent multimodal models.

\textbf{Multi-Violation Amplification Effect:} Counter-intuitively, several approaches demonstrate \emph{improved} performance on \physicsbenchmulti compared to single-violation scenarios. PhyGenEval shows a remarkable 9.3 percentage point accuracy increase on multi-violation cases. This suggests that concurrent violations may create more pronounced deviations from expected behavior, generating stronger detection signals. However, this benefit is exclusive to physics-aware approaches—VideoScore and general anomaly detection methods show degraded performance, underscoring that only models with genuine physics understanding can leverage the additional information in complex scenarios.

\textbf{Dataset Influence:} The performance variation in MIST-VAD variants trained on different datasets (ShanghaiTech vs. UCF-Crime) highlights the importance of training data. UCF-trained variants consistently outperform ShanghaiTech-trained ones, suggesting that UCF-Crime anomalies align better with physics violations.

\textbf{Feature Representation Impact:} The choice of feature representation (C3D vs. I3D) significantly affects performance, with inconsistent patterns across datasets, highlighting the complex interaction between feature extractors and dataset characteristics for physics violation detection.

\textbf{Contextual Understanding:} Approaches incorporating video content descriptions generally outperform those relying solely on physics rule statements, highlighting the importance of contextualizing physical principles within the specific scenario. This effect becomes even more pronounced in multi-violation scenarios, where contextual variants show substantial improvements (e.g., PhygenEval-Contextual-VI: 68.1\% → 80.8\% accuracy) while specific variants show minimal gains or degradation, indicating that holistic scene understanding becomes critical when multiple physics principles interact.

\subsection{Limitations and Challenges}
\label{sec:limitations}

Despite promising results, several challenges remain in effective \physicsbug detection:
    \textbf{Subtle Violations Detection:} Most approaches struggle with subtle physics violations that don't manifest as obvious anomalies, as evidenced by the moderate AUC values (mostly 50-70\%) across approaches.

    \textbf{Context-Dependent Physics:} Our analysis of distribution overlaps suggests that context-dependent physics rules (where the same motion might be valid or invalid depending on the context) remain challenging for all approaches.

    \textbf{Explainability Gaps:} While some approaches provide scores across different dimensions, they cannot generally explain which specific physical laws are being violated and how, limiting their utility for debugging purposes.

    \textbf{Calibration Issues:} The score distributions reveal calibration problems in many approaches, with scores not consistently reflecting confidence levels in physics violation judgments, complicating threshold setting for practical deployment.

In summary, while current approaches show promising capabilities for \physicsbug detection, substantial improvement opportunities remain, particularly in developing methods that combine the semantic understanding strengths of LMMs with the motion pattern detection capabilities of traditional video anomaly approaches, while providing better explainability and calibration.

%% file: sections/results_developer_study.tex
\section{RQ3: Developer Study}

\subsection{Study Design and Participant Demographics}
\label{sec:study_design}

For this research question, we conducted a questionnaire-based study\footnote{The questionnaire and all responses can be found on our website.} with 32 software practitioners to understand their perspectives on detecting \physicsbugs in PE-based systems. The questionnaire covered four main areas: (1) challenges in \physicsbug detection, (2) current practices and tool usage, (3) code-based versus runtime detection, and (4) future needs and expectations.

The majority of respondents (65.6\%) had 0--2 years of experience, while 31.3\% had 3--5 years of experience. Most participants (84.4\%) were involved in development activities, with 50\% also participating in design tasks. Notably, only 18.8\% reported being familiar with physics engines, while another 25\% had some understanding, indicating that physics-based simulation remains a specialized domain.

\subsection{Challenges in Detecting \physicsbugs}
\label{subsec:challenges}

Developers rated the challenge of detecting \physicsbugs as moderate to high (average rating: 3.7/5). 
The results show the primary challenges identified by participants, where the most common difficulties were:

\begin{itemize}[leftmargin=*]
    \item \textbf{Subtle nature of physics errors:} 87.5\% of developers reported that \physicsbugs are difficult to notice with the naked eye, making detection challenging.
    \item \textbf{Unpredictable runtime behavior:} 75\% identified the dynamic nature of physics simulations as a significant obstacle.
    \item \textbf{Sparse occurrence in long videos:} 68.8\% mentioned that \physicsbugs often appear briefly in lengthy runtime sessions, making detection time-consuming.
    \item \textbf{Lack of code-level indicators:} 62.5\% noted that \physicsbugs are not explicitly revealed in code, complicating traditional debugging approaches.
\end{itemize}

When asked about the critical nature of subtle physics failures, 78.1\% of developers considered them important for software reliability. One participant said: ``Even minor physics errors can accumulate and lead to larger problems over time---it's the butterfly effect in action.'' Another noted: ``These subtle failures significantly impact user experience, especially in immersive applications like VR.''

\subsection{Current Practices and Tool Effectiveness}
\label{subsec:current_practices}

Our study revealed a heavy reliance on manual detection methods. 
Results show that 84.4\% of developers rely on visual inspection at least sometimes for identifying \physicsbugs, despite rating its effectiveness as only moderate (average: 3.1/5). Notably, very few developers (9.4\%) were familiar with specialized automated tools for detecting \physicsbugs.

When evaluating current detection approaches, only 12.5\% of participants found them satisfactory, while 53.1\% reported partial satisfaction. Table X summarizes developer perspectives on the limitations of current detection methods:

\begin{table}[h!]
    \centering
    \caption{Limitations of current detection methods}
    \begin{tabular}{|l|c|}
        \hline
        \textbf{Limitation} & \textbf{Percentage} \\
        \hline
        Insufficient understanding of physics laws & 81.3\% \\
        Lack of training data for complex behaviors & 78.1\% \\
        Inability to handle long videos or multiple bugs & 78.1\% \\
        Poor integration into existing workflows & 75.0\% \\
        \hline
    \end{tabular}
    \label{tab:limitations}
\end{table}

One developer commented: ``Current tools don't understand the nuances of physical interactions. They can spot obvious glitches but miss subtle violations that might be critical in certain contexts.''

\subsection{Perceived Importance and Future Needs}
\label{subsec:future_needs}

Despite the limitations of current approaches, developers expressed strong interest in automated \physicsbug detection. Most respondents (75\%) wanted to see such capabilities integrated into existing testing tools and CI/CD pipelines, with an average importance rating of 3.6/5.

Regarding detection approaches, 56.3\% of developers believed that both static code analysis and runtime behavior analysis are equally important for effective \physicsbug detection. However, when asked specifically about runtime-only detectable bugs, participants rated the criticality of automation as 3.9/5, suggesting a strong need for runtime analysis tools.

Results illustrate developers' desired features for future \physicsbug detection tools, where the most requested capabilities were:

\begin{itemize}[leftmargin=*]
    \item Real-time detection during simulations (81.3\%)
    \item Visualization tools to highlight \physicsbugs (78.1\%)
    \item High accuracy in detecting subtle failures (75.0\%)
    \item Integration with existing testing frameworks (68.8\%)
    \item Support for multi-error detection and categorization (65.6\%)
\end{itemize}

Overall, developers rated the potential value of automated \physicsbug detection tools highly (4.0/5), indicating strong support for research in this area.

\subsection{Summary of Key Findings}
\label{subsec:summary}

Our developer study reveals a significant gap between the importance of \physicsbug detection and the effectiveness of current approaches. While developers consider \physicsbugs critical to software reliability, they largely rely on manual methods with moderate effectiveness. The results underscore four key needs:

\begin{itemize}[leftmargin=*]
    \item Tools that can detect subtle physics violations that are easily missed by human inspection
    \item Integration with existing development workflows to reduce the manual effort currently required
    \item Approaches that combine code-level and runtime behavior analysis
    \item Support for detecting multiple coexisting bugs in complex runtime scenarios
\end{itemize}

These findings align with our technical evaluation results from RQ2, confirming that current detection approaches require significant improvements to meet developer needs in real-world scenarios.

%% file: sections/discussion.tex
\section{Threats to Validity}

\subsection{Internal Validity}
The primary internal threat stems from potential biases in our dataset curation process. Although we employed a systematic approach involving multiple annotators and expert validation, subjective judgments may have influenced the classification of \physicsbugs. To mitigate this threat, we implemented a multi-stage review process and consulted domain experts in physics for validation. Additionally, our experiments with detection approaches might be affected by parameter configurations, as different settings could yield varying performance. We addressed this by using standardized configurations and reporting comprehensive metrics across multiple runs.

\subsection{External Validity}
Regarding generalizability, our study is limited by the scope of the collected dataset. While PhysicsBench includes diverse physics failures across various applications, it may not capture all possible manifestations of \physicsbugs in PE-based systems. The taxonomy we developed might, therefore, not be exhaustive. 
Additionally, our developer study involved 32 participants, predominantly with 0-2 years of experience, which may not represent the perspectives of highly experienced practitioners or those in specialized domains. We attempted to mitigate this by recruiting participants from diverse software development roles and backgrounds.

%% file: sections/related-work.tex
\section{Related Work}
\label{sec:related-work}

\subsection{Physics Engine Testing}

Physics engines (PEs) are integral to a wide range of applications, from game development and augmented reality (AR) to robotics and autonomous systems. Testing these engines for correctness and robustness is a crucial challenge, given the complexity of simulating realistic physical interactions. Recent research efforts have focused on various dimensions of PE testing, including fuzzing, reinforcement learning-based exploration, and automated oracle prediction.

Bergdahl et al. introduced a deep reinforcement learning-based framework for automated game testing, which enhances coverage and uncovers logical errors by autonomously exploring complex environments using reward signals~\cite{bergdahl2020augmenting}. Raf et al. proposed PredART, a deep learning-based oracle framework that combines convolutional neural networks (CNNs) and multi-layer perceptrons (MLPs) to assess the realism of virtual object placements in AR applications~\cite{rafi2022predart}. More recently, Xiao et al. developed PHYFU, a fuzzing framework tailored for modern physics engines. By generating perturbation-based inputs, PHYFU uncovers violations of physical laws, offering a systematic approach to identifying faults in physics simulations~\cite{xiao2023phyfu}.

These works underscore the growing emphasis on advancing PE testing methodologies. However, existing approaches often focus on isolated testing techniques and lack comprehensive strategies to address semantically intricate physics failures that manifest as dynamic, runtime deviations from expected behavior.

\subsection{Automated 3D GUI Testing Oracles}

The rise of graphically rich and interactive software systems has driven the need for automated oracles capable of detecting bugs in 3D graphical user interfaces (GUIs) and gameplay environments~\cite{li2020exploratory, li2023towards, li2024grounded, li2024xrzoo, li2025extended, li20253dsoftwaresynthesisguided}. State-of-the-art research in this area has leveraged computer vision techniques, multimodal models, and reinforcement learning for visual bug detection.

Chen et al. proposed GLIB, a framework that utilizes convolutional neural networks for detecting and localizing UI glitches in video games, supported by synthetic datasets generated through code injection and data augmentation~\cite{paper:glib}. Macklon et al. focused on detecting visual inconsistencies in HTML5 canvas-based games by leveraging automatically generated oracle assets and comparing them with rendered images using similarity metrics~\cite{macklon2022automatically}. Taesiri et al. extended this line of research by employing the CLIP model to detect gameplay bugs in videos through zero-shot transfer learning, enabling automated identification of semantic inconsistencies without explicit training on game-specific data~\cite{taesiri2022clip}.

More recent efforts have explored multimodal approaches for anomaly detection in gameplay videos and GUIs. Meng et al. proposed GlitchBench, a benchmarking dataset and evaluation framework for assessing the ability of large multimodal models (LMMs) to detect glitches in video games~\cite{taesiri2024glitchbench}. Similarly, research by Li et al. investigated stereoscopic inconsistencies in VR applications, which are a key cause of cybersickness, highlighting the need for automated detection techniques tailored for extended reality systems~\cite{li2024less}.

These works reveal promising directions for integrating machine learning and multimodal analysis into 3D GUI testing. However, their applicability to runtime physics failure detection remains limited, as they often focus on visual artifacts rather than the underlying physics behaviors.

\subsection{Video-based Testing and Evaluation}

The evaluation of physical consistency in videos, including those generated by text-to-video models, has gained attention in recent years. Such research aligns with our goal of detecting physics bugs by assessing whether simulated behaviors adhere to real-world physical laws.

Meng et al. proposed PhyGenEval, a hierarchical evaluation framework that uses GPT-4 or multimodal models like LLaVA to perform frame-by-frame physical phenomenon detection and sequence-level order verification~\cite{meng2024towards}. Liao et al. introduced Devil, which assesses the naturalness of generated videos through prompt-based scoring of realism using advanced vision-language models~\cite{liao2024evaluation}. Complementarily, Bansa et al. developed VIDEOCON-PHYSICS, a fine-tuned video-text evaluation model trained on annotated datasets for assessing semantic and physical coherence in generated videos~\cite{bansal2024videophy}.

These approaches highlight the potential of leveraging video-based evaluations for detecting physical inconsistencies. However, their primary focus on synthetic video generation differs from our work, which targets runtime failures in PE-based software systems.

%% file: sections/conclusion.tex
\section{Conclusion}
\label{sec:conclusion}

In this study, we addressed the critical challenge of detecting physics bugs in physics engine-based systems, a problem that threatens the reliability of diverse modern applications. We presented the first comprehensive evaluation of cutting-edge techniques, analyzing their effectiveness in identifying physics bugs from runtime behaviors. Our empirical investigation spanned a range of methodologies, including advanced deep learning-based anomaly detection, prompt-based reasoning with large multimodal models, and tailored multimodal learning strategies. By curating a specialized dataset of both faulty and correct simulations, we provided a robust benchmarking framework, guiding future research towards more reliable and effective physics bug detection methods.

%% file: sections/developer_study.tex
\section{Developer Study Questionnaire}

This section shows the developer study questionnaire. The questionnaire is designed to gather insights from developers regarding the importance of detecting physics failures in software systems based on runtime behaviors and their perceptions of the current physics failure detection approaches.

\subsection{Participant Background}

\begin{enumerate}
    \item What roles have you taken in sofwtare / system development process? (Support multiple selections)  
    \begin{itemize}
        \item Design
        \item Development
        \item Quality assurance / Testing
        \item Deployment
        \item Maintainence
        \item Other: \_\_\_\_\_\_\_
    \end{itemize}

    \item How many years of experience do you have in software development process?  
    \begin{itemize}
        \item 0--2 years
        \item 3--5 years
        \item 6--10 years
        \item 10+ years
    \end{itemize}

    \item Are you familiar with physics engines or physics-based simulations? [Explanation: Physics engines refer to software or frameworks used to simulate physical systems and interactions, such as forces, collisions, and motion, in a virtual environment. Examples include Unity Physics, Unreal Engine Physics, Bullet Physics, or custom physics simulations in programming.]
    \begin{itemize}
        \item Yes
        \item No
        \item Somewhat
    \end{itemize}

    \item Have you worked on any projects involving physics engines (e.g., game development, robotics, VR/AR, simulation software)?  
    \begin{itemize}
        \item Yes
        \item No
        \item If yes, please briefly describe: \_\_\_\_\_\_\_
    \end{itemize}
\end{enumerate}

\subsection*{Challenges in Physics Failure Detection}

\begin{enumerate}[resume]
    \item How challenging do you find it to detect physics-related bugs during the development or testing process?  
    (1 = Not Challenging, 5 = Extremely Challenging)  
    \item Which of the following makes physics failure detection difficult? (Select all that apply)
    \begin{itemize}
        \item Physics bugs are subtle and hard to notice with the naked eye.
        \item Errors occur sparsely in long videos, making detection time-consuming.
        \item Runtime behavior is dynamic and hard to predict.
        \item The code does not explicitly reveal physics bugs.
        \item Existing tools lack support for detecting physics-related failures.
        \item Other: \_\_\_\_\_\_\_
    \end{itemize}

    \item In your opinion, how often do physics bugs occur in systems that use physics engines?  
    \begin{itemize}
        \item Rarely
        \item Occasionally
        \item Frequently
        \item Very Frequently
    \end{itemize}

    \item Do you think subtle physics failures (e.g., delayed gravity, imperceptible force errors) are critical for software reliability? Why or why not?  
    \_\_\_\_\_\_\_\_\_\_\_\_\_\_\_\_\_\_\_\_\_\_\_\_\_\_\_\_\_\_\_\_\_\_\_\_\_\_\_\_\_\_\_\_\_
    
    \item How often do you rely on visual inspection (manual video review) to detect physics bugs?  
    \begin{itemize}
        \item Never
        \item Rarely
        \item Sometimes
        \item Frequently
        \item Always
    \end{itemize}

    \item How effective do you find visual inspection for detecting subtle physics failures?  
    (1 = Not Effective, 5 = Extremely Effective)
\end{enumerate}

\subsection*{Current Approaches and Automation}

\begin{enumerate}[resume]
    \item Are you familiar with any automated tools or techniques for detecting physics bugs (e.g., anomaly detection, deep learning-based tools)?  
    \begin{itemize}
        \item Yes
        \item No
        \item If yes, please list the tools/techniques: \_\_\_\_\_\_\_
    \end{itemize}

    \item How effective do you think current automated tools are for detecting physics bugs?  
    (1 = Not Effective, 5 = Extremely Effective)  

    \item In the results you have seen from current approaches, do you feel their accuracy or reliability is satisfactory?  
    \begin{itemize}
        \item Yes
        \item No
        \item Partially
    \end{itemize}
    Please elaborate: \_\_\_\_\_\_\_\_\_\_\_\_\_\_\_\_\_\_\_\_\_\_

    \item Which of the following issues do you think contribute to the poor performance of current automated approaches? (Select all that apply)
    \begin{itemize}
        \item Insufficient understanding of physics laws by the tools.
        \item Lack of training data for complex runtime behaviors.
        \item Tools cannot handle long videos or multiple bugs simultaneously.
        \item Tools are not integrated well into existing development workflows.
        \item Other: \_\_\_\_\_\_\_
    \end{itemize}

    \item Would you like to see automated physics failure detection integrated into existing testing tools or CI/CD pipelines?  
    \begin{itemize}
        \item Yes
        \item No
        \item Maybe
    \end{itemize}
    
    \item If integrated into automated testing, how important do you think physics failure detection would be for your project?  
    (1 = Not Important, 5 = Extremely Important)
\end{enumerate}

\subsection*{Code-Based vs. Runtime Detection}

\begin{enumerate}[resume]
    \item Do you think detecting physics bugs at the \textbf{code level} (e.g., static analysis) is feasible?  
    \begin{itemize}
        \item Yes
        \item No
        \item Maybe
    \end{itemize}
    Please explain why or why not: \_\_\_\_\_\_\_\_\_\_\_\_\_\_\_\_\_\_

    \item Which do you think is more effective for detecting physics bugs:  
    \begin{itemize}
        \item Static code analysis.
        \item Runtime behavior analysis (e.g., video evaluation, simulations).
        \item Both approaches are equally important.
    \end{itemize}

    \item If physics failures are only detectable at runtime, how critical is it to automate their detection?  
    (1 = Not Critical, 5 = Extremely Critical)

    \item Do you believe tools combining both code-level analysis and runtime behavior analysis would improve physics failure detection?  
    \begin{itemize}
        \item Yes
        \item No
        \item Maybe
    \end{itemize}
\end{enumerate}

\subsection*{Future Needs and Expectations}

\begin{enumerate}[resume]
    \item What features would you like to see in an ideal physics bug detection tool? (Select all that apply)
    \begin{itemize}
        \item Real-time bug detection during simulations.
        \item High accuracy in detecting subtle failures.
        \item Visualization tools to highlight physics bugs in video outputs.
        \item Integration with existing testing frameworks (e.g., CI/CD).
        \item Support for multi-bug detection and categorization.
        \item Customizable rules for detecting domain-specific physics violations.
        \item Other: \_\_\_\_\_\_\_
    \end{itemize}

    \item On a scale of 1 to 5, how valuable do you think automated physics bug detection tools would be for improving software reliability?  
    (1 = Not Valuable, 5 = Extremely Valuable)

    \item Any additional comments or thoughts on the importance of detecting physics failures in software systems?  
    \_\_\_\_\_\_\_\_\_\_\_\_\_\_\_\_\_\_\_\_\_\_\_\_\_\_\_\_\_\_\_\_\_\_\_\_\_
\end{enumerate}